\documentclass[letterpaper, 10 pt, conference]{ieeeconf} 
\IEEEoverridecommandlockouts 
\usepackage{graphics} 
\usepackage{times} 
\usepackage{amsmath} 
\usepackage{amssymb} 
\usepackage{subfigure}
\usepackage{multirow} 
\usepackage{amsfonts}
\usepackage{color,soul}
\usepackage{textcomp}
\usepackage{wasysym}
\usepackage{url}
\usepackage{cite}
\usepackage[mathscr]{eucal}
\usepackage{siunitx}
\usepackage{epstopdf}
\usepackage{verbatim}
\usepackage{arydshln}
\usepackage{multirow}
\usepackage{siunitx}
\usepackage{url}

\usepackage[utf8]{inputenc}
\usepackage[english]{babel}
\usepackage{gensymb}
\usepackage{lipsum} 
\usepackage{xargs} 
\usepackage{mathtools}
\usepackage{graphicx}
\makeatletter
\let\NAT@parse\undefined
\makeatother
\usepackage{lipsum} 
\usepackage{xargs} 
\usepackage[pdftex,dvipsnames]{xcolor} 
\usepackage[colorinlistoftodos,prependcaption,textsize=tiny]{todonotes}
\newcommandx{\unsure}[2][1=]{\todo[linecolor=red,backgroundcolor=red!25,bordercolor=red,#1]{#2}}
\newcommandx{\change}[2][1=]{\todo[linecolor=blue,backgroundcolor=blue!25,bordercolor=blue,#1]{#2}}
\newcommandx{\info}[2][1=]{\todo[linecolor=OliveGreen,backgroundcolor=OliveGreen!25,bordercolor=OliveGreen,#1]{#2}}
\newcommandx{\improvement}[2][1=]{\todo[linecolor=Plum,backgroundcolor=Plum!25,bordercolor=Plum,#1]{#2}}
\newcommandx{\thiswillnotshow}[2][1=]{\todo[disable,#1]{#2}}
\usepackage[scaled=.65]{beramono}
\usepackage[ruled, vlined,commentsnumbered]{algorithm2e}
\usepackage{mathtools}
\SetKwComment{Comment}{}{}
\usepackage{xcolor}
\SetCommentSty{mycommfont}

\newtheorem{lem}{Lemma}

\newtheorem{Pro}{Problem}

\newcommand{\eqspace}{\ }

\title{\LARGE \bf Online Camera-to-ground Calibration for Autonomous Driving}

\author{Binbin~Li,
\;Xinyu~Du,
\;Yao~Hu,
\;Hao~Yu,
and Wende~Zhang
\thanks{ B.~Li, X.~Du, Y.~Hu, H.~Yu and W.~Zhang are with General Motors, Warren, MI, 48092, USA {\tt\footnotesize \{binbin.li, xinyu.du, yao.hu, hao.yu, wende.zhang\}@gm.com}
}
}

\graphicspath{{figure/}}

\begin{document}

\maketitle
\thispagestyle{empty}
\pagestyle{empty}

\begin{abstract}\label{sec:abstract}
Online camera-to-ground calibration is to generate a non-rigid body transformation between the camera and the road surface in a real-time manner. Existing solutions utilize static calibration, suffering from environmental variations such as tire pressure changes, vehicle loading volume variations, and road surface diversity. Other online solutions exploit the usage of road elements or photometric consistency between overlapping views across images, which require continuous detection of specific targets on the road or assistance with multiple cameras to facilitate calibration. In our work, we propose an online monocular camera-to-ground calibration solution that does not utilize any specific targets while driving. We perform a coarse-to-fine approach for ground feature extraction through wheel odometry and estimate the camera-to-ground calibration parameters through a sliding-window-based factor graph optimization. Considering the non-rigid transformation of camera-to-ground while driving, we provide metrics to quantify calibration performance and stopping criteria to report/broadcast our satisfying calibration results. Extensive experiments using real-world data demonstrate that our algorithm is effective and outperforms state-of-the-art techniques.

\end{abstract}

\section{Introduction}\label{sec:intro}
Modern vehicles are equipped with a variety of cameras to obtain rich semantic information pertaining to the surrounding environments, and unify features in a shared bird's-eye view (BEV) to enable interpretable motion planning tasks.  Camera-to-ground calibration plays a critical role in determining geometry transformation for feature locations between the camera coordinate and the ground coordinate.  It helps to remove perspective distortion from cameras to provide a BEV representation space, and also facilitates the estimation of distance from a camera mounted on the vehicle to locations on the ground surface, which is widely used for advanced driver assistance systems (ADAS) and autonomous driving systems \cite{van2018autonomous}.

\begin{figure}[t]
    \centering
    \subfigure[]{\includegraphics[width = 1.6in, height=1.2in, viewport = 10 10 580 470, clip]{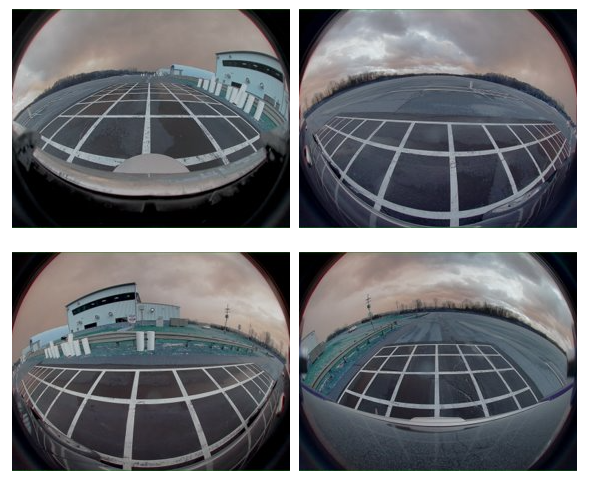}\label{Fig:FEimages}}
    \subfigure[]{\includegraphics[width = 0.6in, height=1.2in, clip]{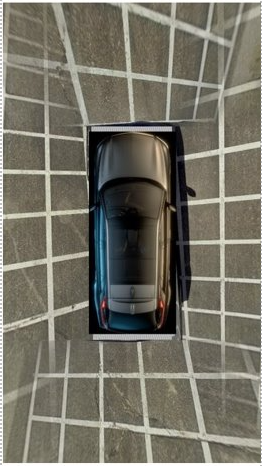}\label{Fig:Bevimage_error}}
    \subfigure[]{\includegraphics[width = 0.6in, height=1.2in, clip]{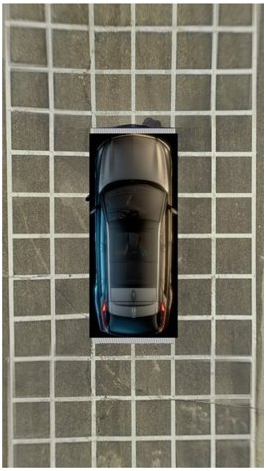}\label{Fig:BEVimage_best}}
    \caption{Given surrounding view fisheye camera images in (a),  static calibration that contains large camera-to-ground estimation errors leads to the mismatched BEV image in (b), but our approach generates a well-aligned BEV image in (c).}\label{Fig:titleFig}
    \vspace{-8mm}
\end{figure}

For the past decades, many methods have been proposed for camera-to-ground calibration. These methods can be generally classified into two categories: (1) static calibration; and (2) online calibration while driving. 
Methods in the first category usually use various patterns like chessboards or manually annotated objects on the ground to calculate camera-to-ground transformation in advance of the driving. However, such a transformation is not rigid because of tire pressure changes, vehicle loading volume variations, road surface diversity, and parts vibration when the vehicle is on the road.  
Camera-to-ground calibration should be repeatedly conducted while driving to adjust geometry projection variations. 
For example, static calibration provides an inaccurate BEV image in Fig. \ref{Fig:Bevimage_error} due to  camera-to-ground  displacement  given images captured by surrounding-view fisheye cameras in Fig. \ref{Fig:FEimages}.
Online calibration mitigates transformation errors, ensures appropriate accommodations for projection variations, and yields the generation of a well-aligned BEV image  in Fig. \ref{Fig:BEVimage_best}.
Existing methods in the second category apply online calibration, which requires specific geometric shapes from the road such as vanishing points from monocular cameras and lane markings extracted from surrounding-view cameras, or photometric consistency between overlapping regions of multiple cameras, to assist calibration adjustment \cite{loukkal2021driving}. 
However, such requirements are difficult to preserve in various  driving environments. 
In this context,  camera-to-ground calibration without relying on any specific targets is required by using continuous images of driving environments from a single camera. 

Instead of utilizing any specific calibration targets,  we perform a coarse-to-fine approach to obtain ground features and optimize for camera-to-ground calibration parameters in a factor graph  when the vehicle is in motion on a roadway. 
We utilize horizon lines to separate ground and non-ground regions in images, predict ground feature locations through wheel odometry, and verify ground features using a geometry-based approach. We perform plane fitting for triangulated ground features to attain the ground normal vectors and the camera center-to-ground height, which are further refined through factor graph optimization to determine camera-to-ground transformation. Considering the non-rigid transformation of camera-to-ground while driving, we also propose metrics to quantify calibration performance, and stopping criteria to ensure the calibration quality. Our algorithm has been demonstrated to be effective using real-world data.

\section{Related Work}\label{sec:RelatedW}

Our research is related to camera-based BEV perception, sensor fusion, and factor graph optimization.

The task of camera-based BEV perception is to unify features from images captured by surrounding cameras into a shared  representation space, which is still very challenging  in the field of low-cost ADAS and autonomous driving.
Typical methods can be classified as \textit{geometry-based methods} and \textit{network-based approaches}. 
\textit{Geometry-based methods} leverage the natural geometric projection to transform camera's perspective view to BEV. 
Can \textit{et al.} \cite{can2022understanding} study scene understanding by online estimation of semantic BEV maps using a single onboard camera.
Loukkal \textit{et al.} \cite{loukkal2021driving} generate BEV occupancy grid maps through a single camera to plan the vehicle motions and provide interpretable intermediate results. 
Ouyang \textit{et al.} \cite{ouyang2020online} propose an extrinsic camera calibration for non-overlapping multi-camera arrays to find the rotation parameters when the vehicle is on a flat horizontal surface.
Recently attention has been drawn to perform BEV fusion through a single camera, multiple cameras, and LiDAR sensor in the learning community. 
Zhu \textit{et al.}  \cite{zhu2021monocular} propose to obtain the BEV image from a single uncalibrated camera without  intrinsic and extrinsic parameters of cameras.
Akan \textit{et al.} \cite{akan2022stretchbev} utilize a stochastic temporal model with BEV representation from multiple cameras to predict the location and motion of all the agents around the ego vehicle. 
More detailed reviews can be found in \cite{ma2022vision}.
Though the aforementioned \textit{network-based methods} are effective to fuse cameras and LiDARs for BEV image generation, they require higher computational resources and extra supports for automotive grade on-board chips. To improve the accuracy and computation speed, we directly extract features from images without relying on any specific targets, and utilize factor graph optimization to obtain the camera-ground calibration parameters on regular commercial  vehicles. 

The use and performance of measurements from multiple sensors directly determine the quantity and quality of information for vehicles with autonomous driving or ADAS.  
Yoo \textit{et al.} \cite{yoo20203d} utilize camera-to-ground transformation from cameras to combine with LiDAR features for object detection. 
Qin \textit{et al.} \cite{qin2020avp} fuse four surround-view cameras to generate the segmentation images on the BEV, and further build a map to facilitate vehicle localization in the parking lot in aid of wheel odometry. 
Song \textit{et al.} \cite{song2017real} present a real-time lane detection and forward collision warning technique in a BEV of a structured environment through camera-to-ground transformation. 
These methods assume static camera-to-ground transformation while driving. Such an assumption does not hold because of the road variations, tire pressure changes, vehicle vibration, and so on. 
In our paper, we use wheel odometer readings from the CAN bus to fuse camera measurements and propose visual-based pipelines to recover the non-rigid camera-to-ground transformation in a real-time manner. 

Factor graph optimization estimates the camera poses at subsequent instants of time through available measurements, which has been approved to be an effective algorithm for batch simultaneous localization and mapping. 
Fan \textit{et al.} \cite{fan2020majorization} propose majorization minimization methods for distributed factor graph optimization problems to guarantee the convergence of first-order critical points under mild conditions. 
Moreira \textit{et al.} \cite{moreira2021fast} leverage the sparsity of the data to allow for high scalability, low computational cost, and high precision by combining the Krylov-Schur method for spectral decomposition with Cholesky factorization. 
Li \textit{et al.} \cite{li2019pose} propose a hybrid visual odometry system to combine an unsupervised monocular visual odometry with a factor graph optimization as a back-end to improve the performance and robustness. 
We formulate our camera-to-ground calibration problem into a factor graph optimization problem  to achieve real-time capability, stability, and robustness on long tracks.

\section{Problem Definition}\label{sec:ProblemFormulation}

\begin{figure*}[htb!]
	\centering
	\includegraphics[width=6.4in, viewport = 12 125 950 450, clip ]{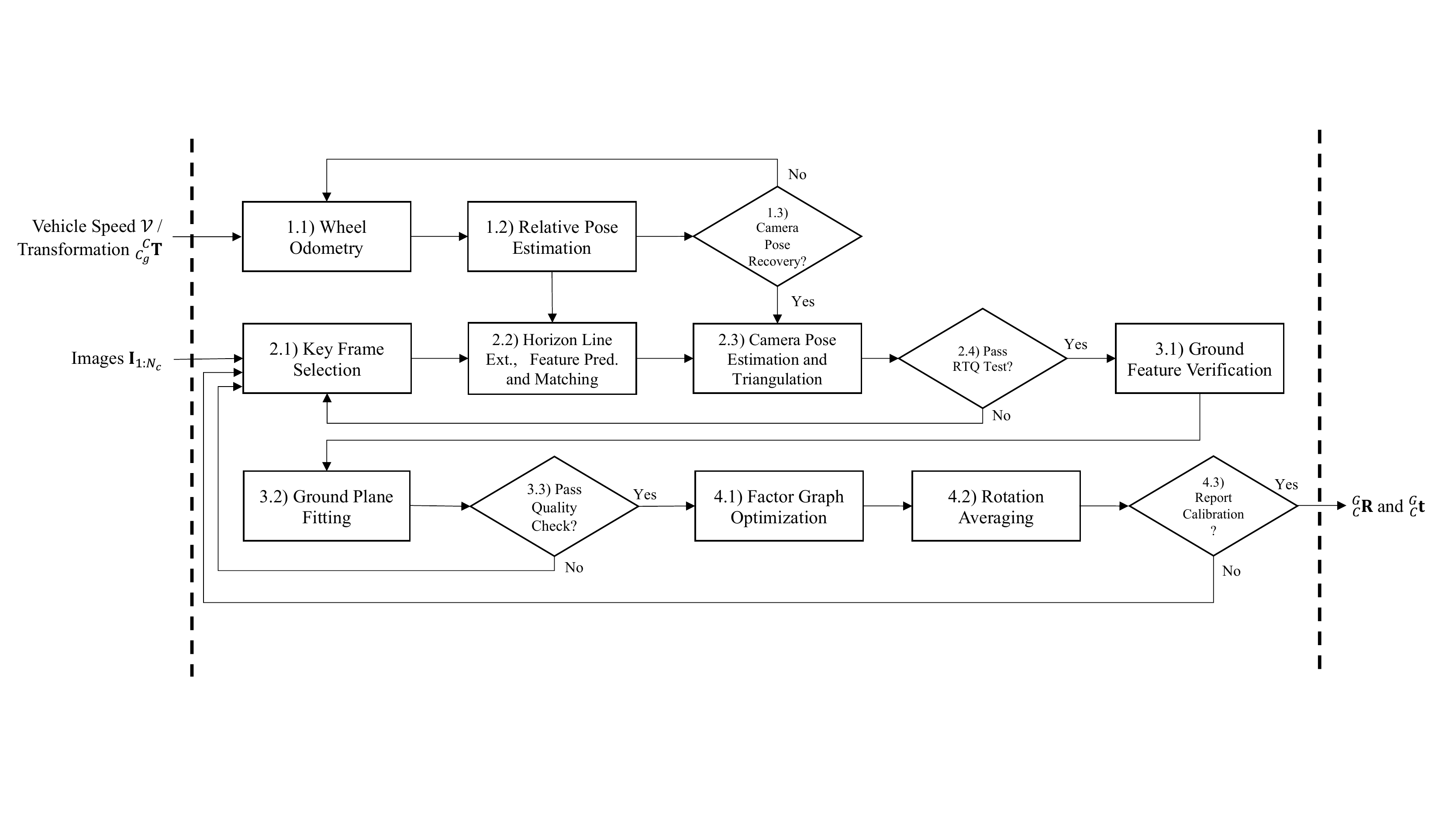}
	\caption{\color{black}System diagram:  Each block is explained in the corresponding subsection of Sec. \ref{sec:SystemModeling} with explicit reference back to this diagram.
	}\label{Fig:SystemDiagram}
        \vspace{-6mm}
\end{figure*}

We assume the vehicle is equipped with a frontal view camera, which is capable to observe the ground surface.  The camera is pre-calibrated, and the nonlinear distortion of images has been removed \cite{zhang2000flexible}.

All coordinate systems are right hand system, and formally defined as follows,
\begin{itemize}
    \item $\{\mathcal{C}\}$ denotes the camera coordinate system with its origin at the camera optical center, its $z$-axis is coinciding with the optical axis and pointing to the forward direction of the camera, and its $x$-axis and $y$-axis are parallel to the horizontal and vertical directions of the camera imaging sensor plane, respectively.
    \item $\{\mathcal{I}\}$ denotes the image coordinate system. Define $\mathbf{p}_{k, j}=[u \; v]^\intercal$ to be the $j^{\text{th}}$ feature position in image $\mathbf{I}_k$, where $(u, v)$ is the image coordinate, $k\in \{1, 2,..., N_c\}$, and $N_c$ is  the number of keyframes. 
    \item $^{\mathcal{Y}}_{\mathcal{X}}\mathbf{T} \in SE(3)$ denotes the transformation matrix from the frame $\{ \mathcal{X}\}$ to $\{\mathcal{Y}\}$.  We also have $^{\mathcal{Y}}_{\mathcal{X}}\mathbf{R} \in SO(3)$ and $ ^{\mathcal{Y}}_{\mathcal{X}}\mathbf{t}$ to be the rotation matrix and translation vector from $\{{\mathcal{X}}\}$ to $\{{\mathcal{Y}}\}$, respectively.
    \item $\{\mathcal{C}_g\}$ denotes the vehicle center of gravity coordinate with its origin at the center of mass, $x$-axis pointing to the vehicle forward moving directions, $y$-axis pointing to the left of the vehicle lateral direction, and $z$-axis pointing upward. Noted that $_{\mathcal{C}_g}^{\mathcal{C}} \mathbf{T}$ can be obtained from manufacturing calibration in advance \cite{lee2018real}.  
    \item $\{\mathcal{G}\}$ denotes the ground coordinate system with its origin right below $\{\mathcal{C}_g\}$'s and located on the ground, $x$-axis pointing vehicle forward moving direction, $y$-axis pointing to the left and parallel to the ground, and $z$-axis pointing upward and perpendicular to the ground plane. 
	\item  $\mathcal{V} = \{({v}_\tau, \delta_\tau) | \tau = 1, 2, ..., N_v\}$ denotes the wheel odometer readings at frame $\{\mathcal{C}_g\}$ from the CAN bus system synchronized with cameras. Here,  $\tau$ is the index when the vehicle speed ${v}_\tau$ is correspondingly generated, $\delta_\tau$ is the front wheel steering angle, and  $N_v$ is the number of samples. 
	\item $\check{\mathbf{X}} = [{\mathbf{X}}^{\intercal}, 1]^{\intercal}$ denotes the homogeneous vector, where ${ \mathbf{X}}$ denotes the inhomogeneous counterpart of $\check{\mathbf{X}}$.
\end{itemize}
Through the pinhole camera model, we have
\begin{equation}\label{eq:groundtocamproj}
	\check {\mathbf{p} }_{k_j} = \lambda \mathbf{K}  
	\left[ 
	\begin{array}{cc}
		{_\mathcal{C}^{\mathcal{G}}}\mathbf{R}^{\intercal} \; |  - {_\mathcal{C}^{\mathcal{G}}}\mathbf{R}^{\intercal} \; {_\mathcal{C}^{\mathcal{G}}}\mathbf{t}
	\end{array}
	\right]_{\textbf{col}: 1, 2, 4}
	{ \check{\mathbf{P} }}_{k, j},
\end{equation}
where $\lambda$ is a scalar, $\mathbf{K}$ is the camera intrinsic matrix,  $()_{\textbf{col}: i}$ means taking the $i^\text{th}$ column of a matrix. Here, we abuse the notation and use $\check{{\mathbf{P}}}_{k, j} = [X, Y, 1]^{\intercal}$ to represent a homogeneous point located on the ground surface in $\{ \mathcal{G} \}$ with $X \in \mathbb{R}$ and $Y \in \mathbb{R}$. 

With the assumptions and notations defined, our problem is defined as follows,
\begin{Pro}
	Given a sequence of image $\mathbf{I}_k$, rigid body transformation $_{\mathcal{C}_g}^{\mathcal{C}} \mathbf{T}$,  and wheel speed $\mathcal{V}$,  obtain ${_\mathcal{C}^{\mathcal{G}}}\mathbf{T}$ while driving. 
\end{Pro}

\section{Methodology}\label{sec:SystemModeling}

Fig. \ref{Fig:SystemDiagram} illustrates our system diagram. It mainly contains the following blocks: (1.1 - 1.3) we utilize the kinematic bicycle model to recover relative motion between image keyframes to facilitate camera pose estimation and ground point triangulation; (2.1 - 2.4) we extract coarse ground features from keyframes, perform feature prediction through vehicle motions,  and further follow a fine ground feature  verification procedure; (3.1 - 3.3) we perform ground  plane fitting obtain the ground normal vector and the camera center-to-ground height; (4.1 - 4.3) we  refine camera poses and camera-to-ground transformation parameters through factor graph optimization, and propose  a stopping criterion that determines when to report/broadcast camera-to-ground calibration. 

\subsection{Camera Motion  via Wheel Odometry}

With consecutive wheel odometer readings from the CAN bus system, we estimate pose changes of the vehicle over time and determine relative motions between camera keyframes to recover scale factors for our monocular camera system. 
Assume the distance from the center of gravity to the rear axles of the vehicle and the distance from the front axles to the rear axles to be $L_r$ and $L$, respectively.  
Denote the world coordinate to be $\{\mathcal{W}\}$, which coincides with   $\{\mathcal{C}_g\}$ at the vehicle starting position.  
We deploy a kinematic bicycle model  and have (see Box 1.1 of Fig. \ref{Fig:SystemDiagram}),
\begin{equation}
	\begin{aligned}\label{eq:vehiclebicyclemodel}
		{^{\mathcal{W}}}\mathbf{P}_{\tau + 1 }  & = {^{\mathcal{W}}}\mathbf{v}_\tau \Delta \tau+ {^{\mathcal{W}}}\dot{\mathbf{v}_\tau} \Delta \tau^2/2 + {^{\mathcal{W}}} \mathbf{P}_{\tau}, \\
		{^{\mathcal{W}}} \mathbf{v}_{\tau + 1} & = {^{\mathcal{W}}} \mathbf{v}_{\tau}  + {^{\mathcal{W}}} \dot{\mathbf{v}}_{\tau}  \Delta \tau, \\
		{_{\mathcal{C}_g}^{\mathcal{W}}} \mathbf{R}_{\tau + 1}   & = {_{\mathcal{C}_g}^{\mathcal{W}}} \mathbf{R}_{\tau} \mbox{Exp} (\omega_\tau \Delta \tau), 
	\end{aligned}
\end{equation}
to represent vehicle position, velocity, and rotation in frame $\{\mathcal{W}\}$  through the wheel speed set $\mathcal{V}$, where  $\Delta \tau$ is the sampling period of the wheel odometer, angular velocity $ \omega_\tau  = [0, 0, v_\tau \sin \beta_\tau /L_r ]^{\intercal}$ with slip angle $\beta_\tau = \tan^{-1} ( L_r \tan \delta_\tau/ L)$,  velocity $ {^{\mathcal{W}}}\mathbf{v}_\tau = [v_x, v_y, 0]^{\intercal}$ with $v_x = v_\tau \cos (\theta_\tau + \beta_\tau)$,  $v_y = v_\tau \sin(\theta_\tau + \beta_\tau)$ and $\dot{\theta}_\tau = v_\tau \sin{\beta_\tau}/L_r$, and $\mbox{Exp}(\cdot)$ represents the exponential map operator \cite{ lee2021novel}.   
Here, ${_{\mathcal{C}_g}^{\mathcal{W}}} \mathbf{R}_{0} =  \mathbf{I}_3$ at the initial position.

It is noted that  the sampling frequency of  the wheel speeds  is higher than that of the cameras, which is common for most modern vehicles.  For consecutive keyframe $\mathbf{I}_k$ and $\mathbf{I}_{k + 1}$, there exist several wheel speeds in time interval $[k, k + 1]$. We iterate the wheel speed integration for all  readings between two consecutive keyframes by (\ref{eq:vehiclebicyclemodel}), and acquire the relative rotation matrix and translation vector of vehicles as 
$ 
\Delta {^{\mathcal{C}_g}} \mathbf{R}_{k}^{k + 1}   = {_{\mathcal{C}_g}^{\mathcal{W}}} \mathbf{R}_{ k + 1}^{\intercal} \: {_{\mathcal{C}_g}^{\mathcal{W}}} \mathbf{R}_{k } = \prod_{\tau = k}^ {k + 1} \mbox{Exp} (\omega_\tau \Delta \tau)
$ 
and 
$
\Delta {^{\mathcal{C}_g}}  \mathbf{P}_{k}^{k + 1}   = \sum_{\tau = k}^{k + 1}  \left[ \mathbf{v}_\tau \Delta_\tau + \frac{1}{2}  {_{\mathcal{C}_g}^{\mathcal{W}}} \mathbf{R}_{\tau}^{\intercal} \dot{\mathbf{v}}_\tau \Delta\tau^2  \right]
$
in frame $\{\mathcal{C}_g\}$, respectively (see Box 1.2 of Fig. \ref{Fig:SystemDiagram}). 
We then recover the rotation matrix and translation vector between keyframe at time $k$ to keyframe at time $k + 1$ in $\{\mathcal{C}\}$ as,
\begin{equation}
	\begin{aligned}\label{eq:relativemotionktok1}
		\Delta {^{\mathcal{C}}} \mathbf{R}_{k}^{k + 1}  &= {_{\mathcal{C}_g}^{\mathcal{C}} {\mathbf{R}}}  \; \Delta {^{\mathcal{C}_g}} \mathbf{R}_{k}^{k + 1} \; {_{\mathcal{C}_g}^{\mathcal{C}} {\mathbf{R}}}^{\intercal}, \\
		\Delta {^{\mathcal{C}}} \mathbf{t}_{k}^{k + 1}  &= 	 {_{\mathcal{C}_g}^{\mathcal{C}} {\mathbf{R}}}  \;   {_{\mathcal{C}_g}^{\mathcal{W}} {\mathbf{R}}_{k + 1}^{\intercal} } ( \mathbf{\Lambda} \; {_{\mathcal{C}_g}^{\mathcal{C}} {\mathbf{R}^{\intercal}}}\;{_{\mathcal{C}_g}^{\mathcal{C}} {\mathbf{t}}}-\Delta {^{\mathcal{C}_g}}  \mathbf{P}_{k}^{k + 1})
	\end{aligned}
\end{equation}
and $\mathbf{\Lambda} = {_{\mathcal{C}_g}^{\mathcal{W}} {\mathbf{R}_{k + 1}}} - {_{\mathcal{C}_g}^{\mathcal{W}} {\mathbf{R}_{k}}}$. 
With relative camera motion  from wheel odometry in (\ref{eq:relativemotionktok1}), we further utilize it for a  coarse to fine ground feature extraction in  consecutive keyframes.

\subsection{Consecutive-keyframe Ground Extraction}

Ground features (features located on the road surface on which the ego vehicle is travelling) play an important role in camera-to-ground calibration. In urban/suburban environment, most ground features are on concrete or asphalt road surfaces, which have similar textures and are difficult to extract and match. We propose a novel coarse-to-fine ground feature extraction architecture for robust camera-to-ground calibration. We first introduce horizon line extraction methods to facilitate feature matching by predicting the locations of ground features through vehicle motion. We then utilize a geometry approach to help verify ground features, and perform ground plane fitting to obtain the ground normal vector and the camera center-to-ground height. 

\subsubsection{Ground Feature Prediction}\label{sec:groundlinefeaturepredic}

We select keyframes to perform calibration while driving at a steady speed (see Box 2.1 of Fig. \ref{Fig:SystemDiagram}).  
Once a keyframe is selected, we start to extract corner features and track them through Kanade–Lucas–Tomasi (KLT) sparse optical flow algorithm \cite{lucas1981iterative}.
Noted the initial value of ${_\mathcal{C}^{\mathcal{G}}} {{\mathbf{T}}}$ can be estimated from ${_\mathcal{C}^{\mathcal{C}_g}} {\mathbf{T}}$ by considering the transformation between $\{\mathcal{C}_g\}$ and $\{\mathcal{G}\}$ through vehicle factory settings.  Otherwise, we use continuous ${_\mathcal{C}^{\mathcal{G}}} {\mathbf{T}}$  from Sec. \ref{sec:optimize}. 
We start with extracting horizon line from images, which defines a visual boundary that separates sky from land or water. 

\begin{figure}[t!]
	\centering
	\includegraphics[width=3 in, viewport = 286 200 600 340, clip ]{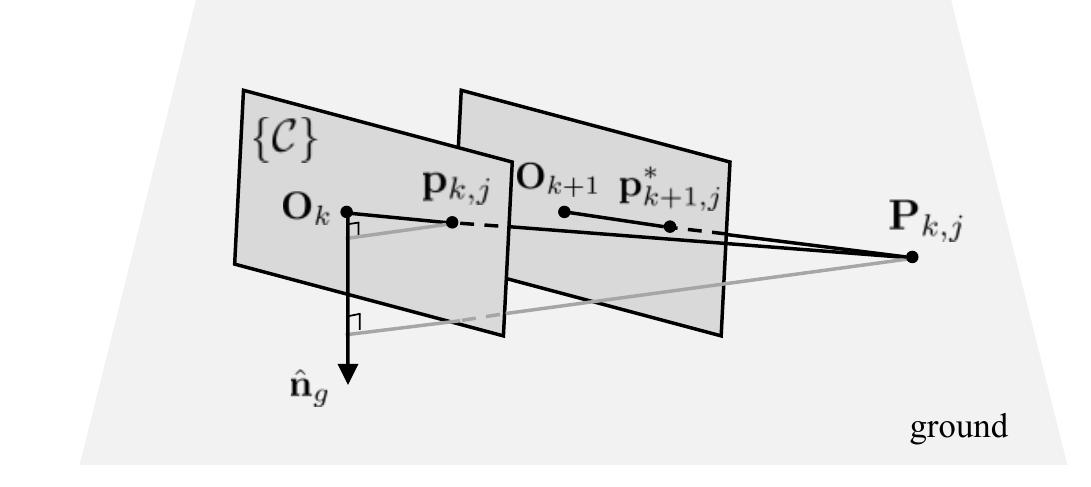}
	\caption{ Feature prediction through vehicle motion. Here, $\mathbf{O}_k$ is the camera center for keyframe $\mathbf{I}_k$.
	}\label{Fig:featurepredictgraph}
        \vspace{-2mm}
\end{figure}

\begin{lem} 
	The horizon line $\mathbf{l}_h$ is identified by two points,
	\begin{itemize}
		\item  ${\check{\mathbf{p}}_1}  = \mathbf{K} [\frac{r_{11}}{r_{31}}, \frac{r_{21}}{r_{31}}, 1 ]^{\intercal}$  and ${\check{\mathbf{p}}_2} = \mathbf{K} [\frac{r_{12}}{r_{32}}, \frac{r_{22}}{r_{32}}, 1 ]^{\intercal}$ if $r_{31} \neq 0$ and $r_{32} \neq 0$;
		\item  ${\check{\mathbf{p}}_1}  = \mathbf{K} [\frac{r_{12}}{r_{32}}, \frac{r_{22}}{r_{32}}, 1 ]^{\intercal}$  and ${\check{\mathbf{p}}_2} = \mathbf{K} [\frac{ r_{11} + r_{12}}{r_{32}}, \frac{r_{21} + r_{22}}{r_{32}}, 1 ]^{\intercal}$ if $r_{31} = 0$ and $r_{32} \neq 0$;
		\item  ${\check{\mathbf{p}}_1}  =  \mathbf{K} [\frac{r_{11}}{r_{31}}, \frac{r_{21}}{r_{31}}, 1 ]^{\intercal}$  and ${\check{\mathbf{p}}_2} = \mathbf{K} [\frac{ r_{11} + r_{12}}{r_{31}}, \frac{r_{21} + r_{22}}{r_{31}}, 1 ]^{\intercal}$ if $r_{31} \neq 0$ and $r_{32} = 0$.
	\end{itemize}
	Here,  $r_{ij}$ is a element of the matrix ${_\mathcal{C}^{\mathcal{G}}} { {\mathbf{R}}}^{\intercal}$ in (\ref{eq:groundtocamproj}), and $i$ and $j$ are the corresponding row and column matrix index, respectively. 
\end{lem}

\begin{proof}
	We discuss four different cases depending on the value of $r_{31}$ and $r_{32}$. 
	\begin{itemize}
	    \item $r_{31} \neq 0$ and $r_{32} \neq 0$: Define ${^{\mathcal{G}}\mathbf{P}}_1 = [1,  \; 0, \; 0]^{\intercal} $ to be a vanishing direction in  $\{\mathcal{G}\}$. We project it back to the unit plane in $\{\mathcal{C}\}$ and have $\tilde{\mathbf{P}}_1 = [\frac{r_{11}}{r_{31}}, \frac{r_{21}}{r_{31}}, 1 ]^{\intercal}$. Obtain its projection in in $\{\mathcal{I}\}$ as ${\check{\mathbf{p}}_1}$ using the pin-hole camera model. Similarly, we have ${\check{\mathbf{p}}_2} = \mathbf{K} [\frac{r_{12}}{r_{32}}, \frac{r_{22}}{r_{32}}, 1 ]^{\intercal}$ for the vanishing direction $^{\mathcal{G}}\mathbf{P}_2 = [0,  \; 1, \; 0]^{\intercal} $. 
	    \item $r_{31} = 0$ and $r_{32} \neq 0$: Project the vanishing direction $^{\mathcal{G}}\mathbf{P}_2 $ back to $\{\mathcal{I}\}$ to  have ${\check{\mathbf{p}}_1} $. Denote a vanishing direction to be $\mathbf{l}_v = [x_v, y_v, 0]^{\intercal}$ in $\{\mathcal{G}\}$, and the corresponding vanishing point to be  $\check{\mathbf{p}}_v = \mathbf{K} [\frac{r_{11}}{r_{32}} \frac{x_v}{y_v} + \frac{r_{12}}{r_{32}}, \frac{r_{21}}{r_{32}} \frac{x_v}{y_v} + \frac{r_{22}}{r_{32}}, 1]^{\intercal}$. Connect the point ${{\mathbf{p}}_1} $ with the point ${\mathbf{p}}_v$ in $\{\mathcal{I}\}$, and we obtain the slope of the horizon line $\mathbf{l}_h$ to be  $s_v = \frac{f_y}{f_x} \frac{r_{21}}{r_{11}}$.  Here, $f_x$  and $f_y$ are camera focal lengths along $x$-axis and $y$-axis of the camera intrinsic $\mathbf{K}$, respectively. 
	    For convenience,  we set $\frac{x_k}{y_k} = 1$ and  obtain  ${\check{\mathbf{p}}_2} = \mathbf{K} [\frac{ r_{11} + r_{12}}{r_{32}}, \frac{r_{21} + r_{22}}{r_{32}}, 1 ]^{\intercal}$.  
	    \item $r_{31} \neq 0$ and  $r_{32} = 0$:  Perform the same operations for the vanishing direction ${^{\mathcal{G}}\mathbf{P}}_1= [ 1,  \; 0, \; 0]^{\intercal} $,  and obtain ${\check{\mathbf{p}}_1}  =  \mathbf{K} [\frac{r_{11}}{r_{31}}, \frac{r_{21}}{r_{31}}, 1 ]^{\intercal}$  and ${\check{\mathbf{p}}_2} = \mathbf{K} [\frac{ r_{11} + r_{12}}{r_{31}}, \frac{r_{21} + r_{22}}{r_{31}}, 1 ]^{\intercal}$. 
	    \item $r_{31} = 0$ and  $r_{32} = 0$: The camera's principal axis points toward the sky. It is against our assumption that the camera can observe the road.  
	\end{itemize}
	With the aforementioned cases presented, the lemma is proved.
\end{proof}

\begin{figure}[tb]
        \centering
	\subfigure[]{\includegraphics[width = 1.4in, height = 0.8in, clip]{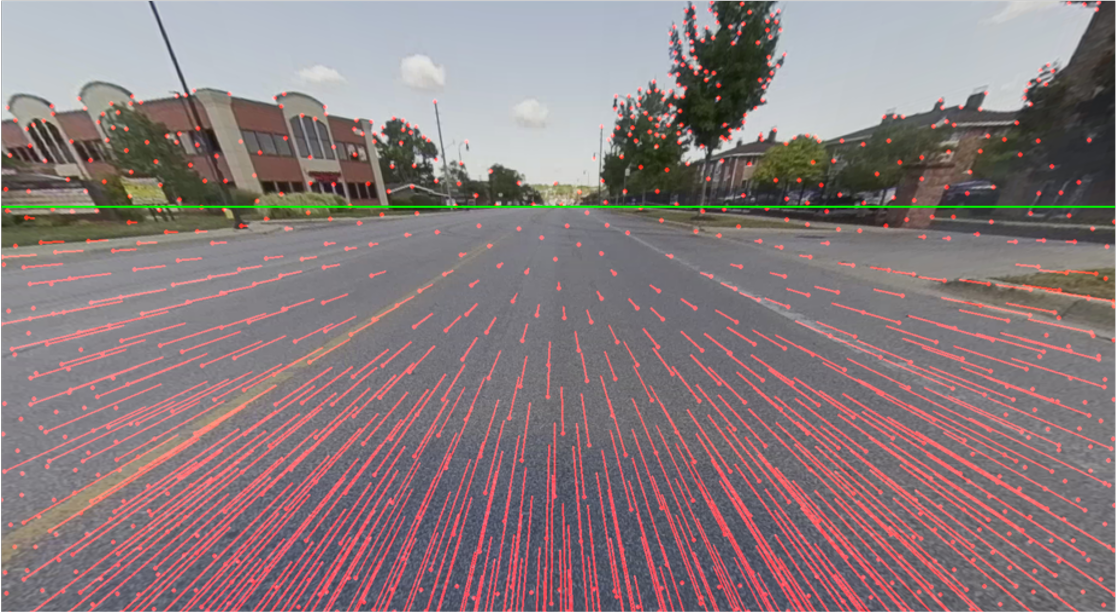}\label{Fig:feature1}}
	\subfigure[]{\includegraphics[width = 1.4in, height = 0.8in, clip]{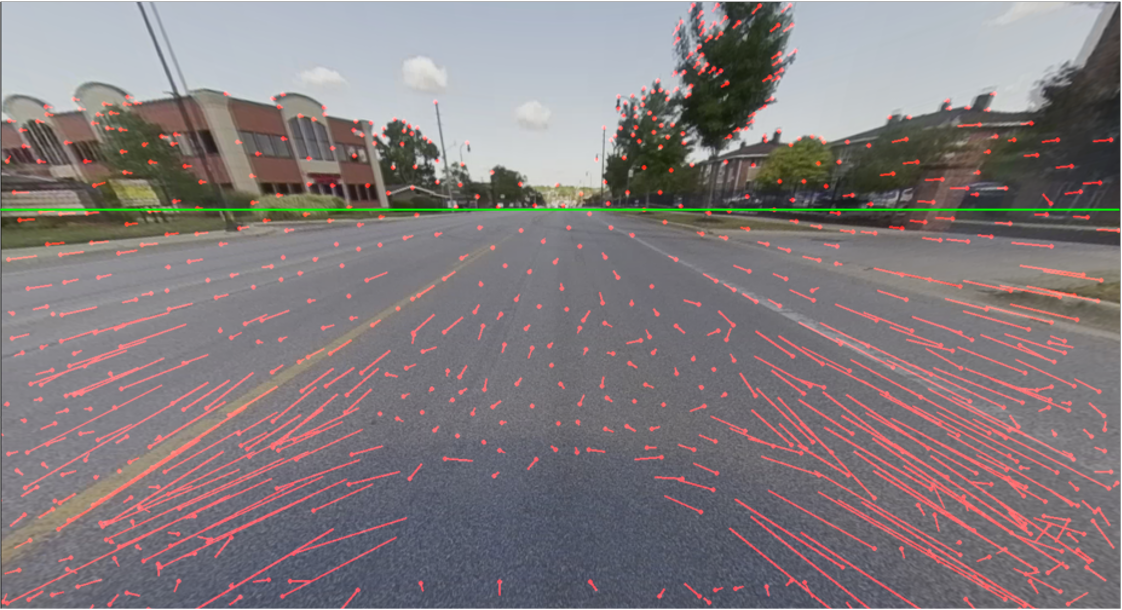}\label{Fig:feature2}}
	\subfigure[]{\includegraphics[width = 1.4in, height = 0.8in, clip]{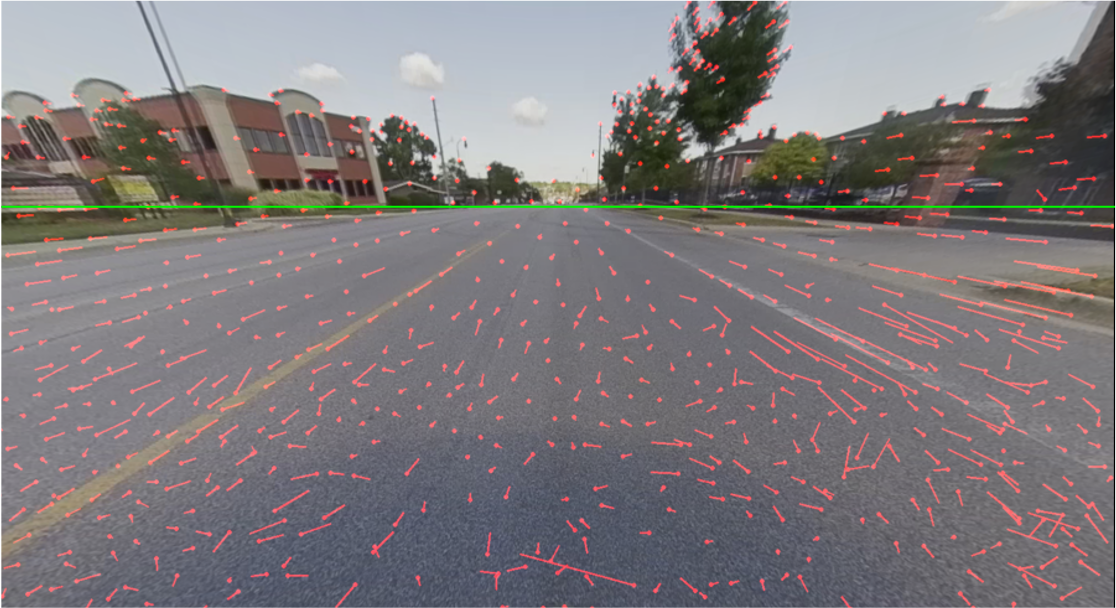}\label{Fig:feature3}}
	\subfigure[]{\includegraphics[width = 1.4in, height = 0.8in,  clip]{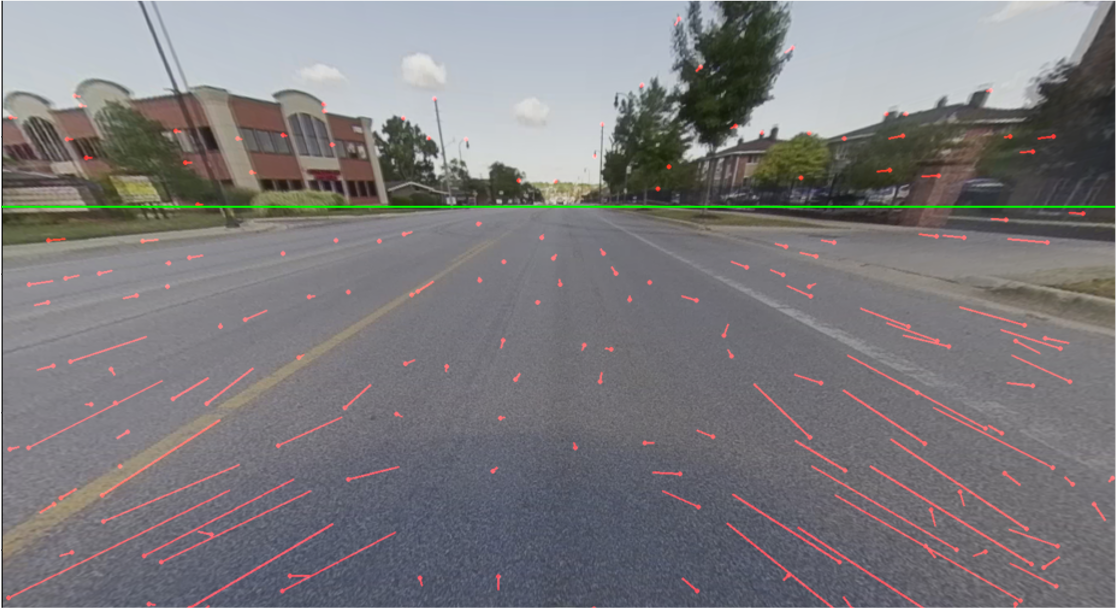}\label{Fig:feature4}}
	\caption{Coarse ground feature extraction.  For each feature, we predict its position in the next keyframe through vehicle motion in (a). Thus we have more considerable and high-quality matched feature pairs in (b) than features from KLT tracker without any predictions in (c). We further evenly sample/select features in (d) from (b) for computational optimization. Here, the endpoints of the red line segments represent matched features between keyframes $\mathbf{I}_k$ and $\mathbf{I}_{k + 1}$, and green lines are the horizon lines (best viewed in color).}\label{Fig:featurepredic}
    \vspace{-3mm}
\end{figure}

The horizon line $\mathbf{l}_h$ separates the image $\mathbf{I}_k$ into two regions and provides a  boundary to determine ground features (see Fig. \ref{Fig:featurepredic}). 
We select coarse ground features with $\mathbf{l}_h^{\intercal} \; \check{\mathbf{p}}_{k, j} > 0$, and further predict ground feature locations in the next keyframe through vehicle motions.

\begin{lem}
	A ground feature $\mathbf{p}_{k, j}$ in image $\mathbf{I}_k$ is  approximated in image  $\mathbf{I}_{k + 1}$ at the position of,
	\begin{align}\label{eq:predFeature}
		{\check{\mathbf{p}}_{k + 1, j}^{\star}} = s_{k+ 1, j} \mathbf{K} ( \frac{ \Delta {^{\mathcal{C}}} \mathbf{R}_{k}^{k + 1} \; \mathbf{K}^{-1}  \; \check{\mathbf{p}}_{k, j} }{ \Vert  {\mathbf{K}^{-1} \check{\mathbf{p}}_{k, j}}  \cdot {\hat{\mathbf{n}}_g} \Vert } + \frac{\Delta {^{\mathcal{C}}} \mathbf{t}_{k}^{k + 1}}{\Vert {_\mathcal{C}^{\mathcal{G}}} {\mathbf{t}} \Vert } ). 
	\end{align}
    Here, $s_{k+ 1, j}$ is a scalar,  $\hat{\mathbf{n}}_g  = {_\mathcal{C}^{\mathcal{G}}} { {\mathbf{R}}_{\mathbf{col}:3}^{\intercal}}$,   $ (\cdot )$ is the vector dot product operator, and $\Vert \cdot \Vert$ is the vector $l^2$-norm. 
\end{lem}

\begin{proof}
    We project the feature $\mathbf{p}_{k, j}$ back to the unit plane at $\tilde{\mathbf{p}}_{k, j} =  \mathbf{K}^{-1} \check{\mathbf{p}}_{k, j}$.
    We express the ground normal vector as $\hat{\mathbf{n}}_g  = {_\mathcal{C}^{\mathcal{G}}} {{\mathbf{R}}}^{\intercal} \; [0, 0, 1]^{\intercal}$ in  $\{\mathcal{C}\}$ (see Fig. \ref{Fig:featurepredictgraph}).
    Define $\mathbf{P}_{k, j}$ to be the 3D point in $\{ {\mathcal{C}} \}$ located on the ground for the corresponding $\tilde{\mathbf{p}}_{k, j}$.
    Through trigonometry, we have
	$
	   \frac{\Vert  \tilde{\mathbf{p}}_{k, j} \Vert}{\Vert \mathbf{P}_{k, j}  \Vert} = \frac{ \Vert \tilde{\mathbf{p}}_{k, j} \cdot  \hat{\mathbf{n}}_g  \Vert }{{ \Vert_\mathcal{C}^{\mathcal{G}}}  {\hat {\mathbf{t}}} \Vert }
	$
	for us to recover the depth for the ground point. 
	With  $\Vert \mathbf{P}_{k, j}  \Vert$ recovered, we obtain ground point $	\mathbf{P}_{k, j}$ in  $\{\mathcal{C}\}$ by
        $
        \mathbf{P}_{k, j}  = \Vert {_\mathcal{C}^{\mathcal{G}}} {\mathbf{t}} \Vert   \tilde{\mathbf{p}}_{k, j} / \Vert  \tilde{\mathbf{p}}_{k, j}  \cdot \hat{\mathbf{n}}_g \Vert.
	$
	We utilize (\ref{eq:relativemotionktok1}) to transform the point $\mathbf{P}_{k, j} $ to  frame $\{\mathcal{C}\}$ at timestamp $k + 1$ by $\mathbf{P}_{k+ 1, j} =  \Delta {^{\mathcal{C}}} \mathbf{R}_{k}^{k + 1} \mathbf{P}_{k, j}   +  \Delta {^{\mathcal{C}}} \mathbf{t}_{k}^{k + 1}$. Project the point $	\mathbf{P}_{k+ 1, j}$ back to the image $\mathbf{I}_{k + 1}$ with the location of the point ${\check{\mathbf{p}}_{k + 1, j}^{\star}}$ (see Fig. \ref{Fig:feature1}), and we have the lemma proved. 
\end{proof}

We apply Lemma 2 to obtain the predicted features located in image $\mathbf{I}_{k + 1}$ for each feature $\mathbf{p}_{k, j}$ in image $\mathbf{I}_k$. The KLT tracker will search around ${{\mathbf{p}}_{k + 1, j}^{\star}}$ for the optimal matched ones, which generates more considerable and high-quality feature pairs (see Fig. \ref{Fig:feature2}).  

\subsubsection{Ground Feature Selection}\label{sec:campose}

We evenly space the image into an equally sized grid,   select the  strongest matched features in each grid, and decrease the total number of features for computational considerations (see Fig. \ref{Fig:feature4}).
We obtain the fundamental matrix $\mathbf{F} = \mathbf{K} ^{-\intercal} \lfloor{^{\mathcal{C}} \mathbf{t}_{k}^{k + 1}} \rfloor_{\times}  {^{\mathcal{C}} \mathbf{R}_{k}^{k + 1}} \mathbf{K}^{-1}$ for   feature pair $\mathbf{p}_{k, j}$ and $\mathbf{p}_{k + 1, j}$ with $\check{\mathbf{p}}_{k, j}^{\intercal} \mathbf{F} \check{\mathbf{p}}_{k + 1, j} = 0$. Here, $\lfloor \mathbf{a} \rfloor_{\times}$ is the matrix representation of the cross product with a vector $\mathbf{a}$, and $\{ {^{\mathcal{C}} \mathbf{R}_{k}^{k + 1}} , {^{\mathcal{C}} \mathbf{t}_{k}^{k + 1}} \}$  is the relative rigid transformation between $\mathbf{I}_{k}$ and $\mathbf{I}_{k + 1}$. Given imprecise estimation of vehicle poses from (\ref{eq:relativemotionktok1}), we consider that $\{ {^{\mathcal{C}} \mathbf{R}_{k}^{k + 1}} , {^{\mathcal{C}} \mathbf{t}_{k}^{k + 1}} \}$  passes our quality check when
$
\angle{({^{\mathcal{C}} \mathbf{R}_{k}^{k + 1}})} \prec \Delta {\theta} \cdot \mathbf{1} \mbox{ and } \Vert {^{\mathcal{C}} \mathbf{t}_{k}^{k + 1}} \cdot \hat{\mathbf{n}}_g\Vert \ge \epsilon_g 
$
to ensure alignment between the vehicle's motion and its heading. 
Here, $ \angle{(\cdot)}$ is the operation that converts rotation matrices to Euler angle representations, $\prec$ denotes the vector component-wise less than operator such that $\mathbf{u}(l) < \mathbf{v}(l)$ ($\forall  l \in \{1, 2,..., n\}$) for vector $\mathbf{u}$ and $\mathbf{v}$, $\mathbf{1}$ presents $3 \times 1$ vector of ones,  $\epsilon_g$ is a scalar, and $\Delta {\theta}$ and $\Delta {t}$ are pre-defined threshold variables, respectively. In practice,  we have $\Delta \theta = 1 \degree$ and $\epsilon_g = 0.95$. 
We re-scale the translation vector ${^{\mathcal{C}} \mathbf{t}_{k}^{k + 1}}$ to be $\Vert \Delta {^{\mathcal{C}}} \mathbf{t}_{k}^{k + 1}   \Vert {^{\mathcal{C}} \mathbf{t}_{k}^{k + 1}}$ (see Box 2.3 of Fig. \ref{Fig:SystemDiagram}), which will be further optimized in Sec. \ref{sec:optimize}.

Define  $\mathcal{P}_{k \leftrightarrow k +1} = \{ (\mathbf{p}_{k, j},  \mathbf{p}_{k + 1, j}) | j = 1, 2,..., N_p\}$ to be the coarse ground features set after fundamental matrix computation \cite{hartley2003multiple}, and $N_p$ is the total number of matched features. Features in set $\mathcal{P}_{k \leftrightarrow k +1} $ may be located on the trees, buildings, cars, and so on. Ground feature verification is proposed to handle the challenge.

\subsubsection{Ground Feature  Verification}

\begin{figure}
	\centering
	\includegraphics[width=2.5in, viewport = {190 88 760 440}, clip ]{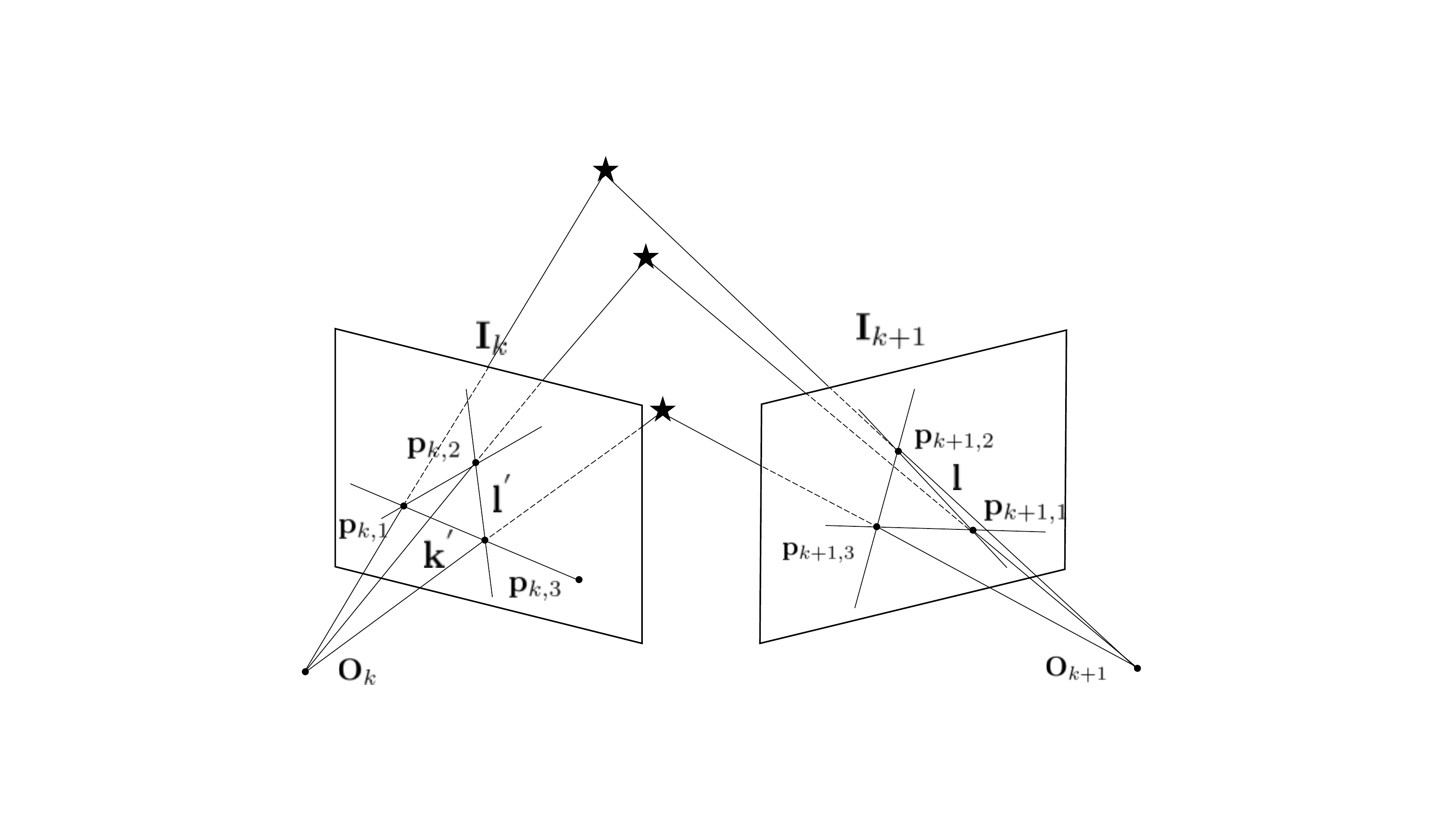}
	\caption{Road feature selection {\color{black} using epipolar geometry.}}\label{Fig:twostepfeaturesdiagram}
\end{figure}

We apply a geometry-based approach to help obtain a fine ground feature set (see Box 3.1 of Fig. \ref{Fig:SystemDiagram}). 
Here, we first derive ${\mathbf{n}_g} \in \mathbb{R}^3$, the referred ground normal vector in $\{ \mathcal{C} \}$ when the keyframe $\mathbf{I}_{k}$ is taken, 
\begin{lem}
The vector ${{\mathbf{n} } }_{g}$ can be obtained from the null space of the following matrix,
$$
\resizebox{1.0\hsize}{!}{$
    \begin{bmatrix}
        \mathcal{F}(\mathbf{p}_{k, 1}, \mathbf{p}_{k, 2}, \mathbf{p}_{k, 3}), \!&\!  \mathcal{F}(\mathbf{p}_{k, 2}, \mathbf{p}_{k, 1}, \mathbf{p}_{k, 3}), \!&\! \mathcal{F}(\mathbf{p}_{k, 1}, \mathbf{p}_{k, 3}, \mathbf{p}_{k, 2})
    \end{bmatrix}^{\intercal}, 
$} 
$$
where  $\{{\mathbf{p}}_{k, l} | l = 1, 2, 3\}$ are three non-collinear points  from the feature set $\mathcal{P}_{k \leftrightarrow k +1} $, ${\mathbf{p}}_{k + 1, l} \in \mathbf{I}_{k + 1}$ is the matched point  for the feature $\mathbf{p}_{k, l}$, and $\mathcal{F}(\mathbf{p}_{k, p}, \mathbf{p}_{k, r}, \mathbf{p}_{k, q}) = ({\check{\mathbf{p}}}_{k, p}\times {\check{\mathbf{p}}}_{k, q}) \times (\mathbf{F} \lfloor{\check{\mathbf{p}}}_{k + 1, p} \times {\check{\mathbf{p}}}_{k + 1, r} \rfloor_{\times} ({\check{\mathbf{p}}}_{k + 1, p} \times {\check{\mathbf{p}}}_{k + 1, q}))$.
\end{lem}
\begin{proof}
	Considering skew matrix properties \cite{trawny2005indirect}, we have $\lfloor{ \mathbf{F}^{-1} \: {\mathbf{n}_g}}\rfloor_{\times} =  \mbox{det}(\mathbf{F})  \; \mathbf{F}^\intercal  {\lfloor{\mathbf{n}_g}\rfloor_{\times} } \; \mathbf{F}$, where $\mbox{det}(\cdot)$ is the determinant of a matrix. Hence we bring in feature $ \mathbf{p}_{k, j}$ on both sides of the equation, and obtain $	{(\mathbf{F} \: \check{\mathbf{p}}_{k, j})^{\intercal}} \:  {\lfloor{\mathbf{n}_g}\rfloor_{\times} } \:  \mathbf{F} \:  \check{\mathbf{p}}_{k, j} = 0$. Define $\mathbf{l}= \mathbf{F} ~ \check{\mathbf{p}}_{k, j}$ to be the epipolar line on image $\mathbf{I}_k$. $\mathbf{l}^{'}$ is the its corresponding epipolar line on image $\mathbf{I}_{k}$, and $\mathbf{k}^{'} $ is a line that does not pass through the epipole of the image $\mathbf{I}_{k}$. We further relate $\mathbf{l}$ and $\mathbf{l}^{'}$ by $\mathbf{l} = \mathbf{F}{\lfloor \mathbf{k}^{'} \rfloor_{\times}} \mathbf{l}^{'}$ (see Fig. \ref{Fig:twostepfeaturesdiagram}). We reorganize the above algebraic equation,   have $\big( \mathbf{l} \times (\mathbf{F} \:  \lfloor {\mathbf{k}^{'} }\rfloor_{\times} \:  \mathbf{l}^{'} ) \big)^{\intercal} \:  {\mathbf{n}_g}= 0$, and lead to $\mathcal{F}(\mathbf{p}_{k, p}, \mathbf{p}_{k, r}, \mathbf{p}_{k, q})^{\intercal} \mathbf{n}_g = 0$,  which helps us obtain the  vector $ {\mathbf{n}_g}$ through three lines. The aforementioned lines can be constructed through image features extracted from set $\mathcal{P}_{k \leftrightarrow k +1} $. Stack for the three non-parallel lines to solve $\mathbf{n}_g$ using singular value decomposition, and we have the lemma approved. 
\end{proof}

Define $L({\mathbf{n}_{g}}, {{\hat {\mathbf{n} } }_{g}}) $ to be the label function  for feature pairs $\mathbf{p}_{k, j}$ and $\mathbf{p}_{k+1, j}$ to identify ground features and we have,  
\begin{equation}\label{eq:normalvector}
	L({\mathbf{n}_{g}}, {{\hat {\mathbf{n} } }_{g}}) = 	\begin{cases}
		1 & \mbox{if } \Vert {\mathbf{n}_{g}} \times {{\hat {\mathbf{n} } }_{g}} \Vert \le \epsilon_l, \\
		0 & \mbox{otherwise.}
	\end{cases}	
\end{equation}
Here, $\epsilon_l$ is a threshold variable that facilitates ground feature selection ($\epsilon_l = 0.2$ in practice).
We label the feature $\mathbf{p}_{k, l}$ as a ground feature if $L({\mathbf{n}_{g}}, {{\hat {\mathbf{n} } }_{g}})$ is equal to $1$; otherwise, at least  one of the three points selected is not located on the ground. 
We sequentially evaluate features in the set $\mathcal{P}_{k \leftrightarrow k+1}$ once three randomly selected feature pairs are ground features, which is executed in linear time.
We then obtain a set of high-quality and refined features as $\mathcal{Q}_{k \leftrightarrow k +1} $ by eliminating ground features with re-projection errors over 1 pixel through triangulation.  
We perform plane fitting for triangulated 3D ground points to obtain the ground normal vector $\mathbf{g}_s$ and camera center-to-ground height $h_s$, which are filtered to remove unrealistic estimations through
$
\Vert \mathbf{g}_s \cdot {^{\mathcal{C}} \mathbf{t}_{k}^{k + 1}}/\Vert {^{\mathcal{C}} \mathbf{t}_{k}^{k + 1}} \Vert   \Vert \ge \epsilon_s \mbox{ and }  \vert h_s - \Vert {_\mathcal{C}^{\mathcal{G}}}  {{\mathbf{t}}} \Vert \vert \le \epsilon_h. 
$
Here, $ \epsilon_s$ and $\epsilon_h$ are threshold variables for ground normal vector and estimated camera center-to-ground height, respectively (we set $\epsilon_s = 0.99$ and $\epsilon_h = 0.05 \; m$ in practice). 

Now we have all the elements to start a sliding window-based  factor graph optimization to refine   camera poses and camera-to-ground transformation. 

\subsection{Cross-keyframe Ground Refinement}\label{sec:optimize}

We utilize a sliding window-based factor graph optimization approach to optimize camera poses,  ground normal vector, and camera center-to-ground height (see Box 4.1 of Fig. \ref{Fig:SystemDiagram}).  
Through the homograph transformation matrix \cite{barfoot2017state},
\begin{align}\label{eq:H}
	\mathbf{H}_k = \mathbf{K}\; {^{\mathcal{C}} \mathbf{R}_{k}^{k + 1}} (\mathbf{I}_{3} + \frac{1}{h_s} {^{\mathcal{C}} \mathbf{t}_{k+1}^{k} }\mathbf{g}_s^{\intercal}) \mathbf{K}^{-1},
\end{align}
we can transfer ground feature $\mathbf{p}_{k, j} \in \mathbf{I}_k$  to the keyframe $ \mathbf{I}_{k + 1}$. 
Given a set of $N_w$ continuous keyframes,  the full state vector in the sliding window is defined as,
\begin{equation}\label{eq:state}
	\begin{aligned}
		\mathcal{X} & = \{ \mathbf{x}_1, \mathbf{x}_2,..., \mathbf{x}_{N_w}, \mathbf{x}_s\}, \\
		\mathbf{x}_k &= \{\angle{({^{\mathcal{C}} \mathbf{R}_{k}^{k + 1}})}, {^{\mathcal{C}} \mathbf{t}_{k}^{k + 1} }\}, k \in \{1,2,..., N_w\}, \\
		\mathbf{x}_s &= \{\mathbf{g}_s, h_s\}.
	\end{aligned}
\end{equation}  
We aim to solve the following minimizing problem utilizing nonlinear solvers in \cite{dellaert2012factor}, 
\begin{equation}\label{eq:optimize}
\resizebox{1.0\hsize}{!}{$
\underset{\mathcal{X}}{\min}  \Big \{ \!  \Vert \mathbf{r}_m - \mathbf{H}_m \mathcal{X}\Vert + 
\sum_{k = 1}^{N_w} \sum_{j = 1}^{\vert \mathcal{Q}_{k \leftrightarrow k +1} \vert} \rho ( \Vert \underbrace{\check{\mathbf{p}}_{k + 1, j}  \! - \!   \lambda {\mathbf{H}_k }  \check{\mathbf{p}}_{k, j}}_{\mathbf{r}_{k, j}} \Vert_{\mathbf{P}_{k, j}})\Big \},  
 $}
\end{equation}
\vspace{-11mm}
\begin{flalign*}
& s.t. { \: \: \Vert \mathbf{g}_s \Vert = 1, {^{\mathcal{C}} \mathbf{R}_{k}^{k + 1}} {^{\mathcal{C}} \mathbf{R}_{k + 1}^{k}} = \mathbf{I}_{3}, }&
\end{flalign*}
where $\lambda$ is a scalar, $\mathbf{P}_{k, j}$ is the standard covariance of a fixed length in the tangent space, $\rho (\cdot)$ is the Huber norm, 
and $\{ \mathbf{r}_m, \mathbf{H}_m \}$ is the prior information for marginalization \cite{li2019pose}. Here, we incorporate marginalization in order to bound the computational complexity of our optimization-based system. 

\begin{table}[t]
	\caption{ Testing Dataset } 
	\centering 
        \addtolength{\tabcolsep}{-4pt}    
	\begin{tabular}[htb!]{ c c c c c }
		\hline
		\\\\[-2.5\medskipamount]
		Sequence & Duration ($s$) & length ($m$) & Weather & \% of Driving\\ 
		\\\\[-2.8\medskipamount]
		\hline	 		
		\\\\[-2.8\medskipamount]
		{FPG}                 & 309   & 2086  & Cloudy              & 91\% \\ 
		{City I (Daytime)}    & 948  & 6577   & Sunny               & 22\% \\ 
            {City II (Nighttime)} & 926  & 13087  & Partly Cloudy       & 34\% \\ 
            {Urban}               & 310  & 2087   & Partly Sunny        & 91\% \\ 
            {Suburban}            & 1084  & 4639  & Rainy               & 42\% \\ 
		{Rural}               & 438   & 4693  & Clear Sky           & 14\% \\ 
		\\\\[-2.5\medskipamount] 
		\hline
	\end{tabular} \label{tab::datainfo}
        \addtolength{\tabcolsep}{4pt}    
	\vspace{-5mm}
\end{table}

We use  vector $\mathbf{g}_s^{*}$ to recover the camera-to-ground rotation, where $(*)$ represents optimized values in $\mathcal{X}$ from (\ref{eq:optimize}). We apply Gram–Schmidt process to  orthonormalize  vector $\mathbf{g}_s^{*}$ and ${^{\mathcal{C}}} \mathbf{t}_{*, k}^{k + 1}$ in an inner product space, and have  $\mathbf{n}_x = {{^{\mathcal{C}}} \mathbf{t}_{*, k}^{k + 1}}/{\Vert {^{\mathcal{C}}} \mathbf{t}_{*, k}^{k + 1} \Vert}$ and  $\mathbf{n}_z = \mathbf{g}_s^{*}- {(\mathbf{g}_s^{*} \cdot \mathbf{n}_x)\mathbf{n}_x}$. The orthogonal matrix ${_\mathcal{C}^{\mathcal{G}}}\mathbf{R}_{k}^{*}$ can be constructed by,
\begin{equation}\label{eq:finalRot}
    {_\mathcal{C}^{\mathcal{G}}}\mathbf{R}_{k}^{*}  =  
    \begin{bmatrix}
        \mathbf{n}_x, \;\; \lfloor \mathbf{n}_x \rfloor_ \times \frac{\mathbf{n}_z}{\Vert \mathbf{n}_z \Vert}, \;\; \frac{\mathbf{n}_z}{\Vert \mathbf{n}_z \Vert}
    \end{bmatrix}^{\intercal}.
\end{equation}
Through  (\ref{eq:optimize}) and (\ref{eq:finalRot}), we obtain the optimized camera-to-ground rotation matrix ${_\mathcal{C}^{\mathcal{G}}}\mathbf{R}_{k}^{*}$, and the corresponding translation vector ${_\mathcal{C}^{\mathcal{G}}}\mathbf{t}_{k}^{*}$ by replacing   ${_\mathcal{C}^{\mathcal{G}}} {{\mathbf{t}}}_{3}$ from factory settings with $h_s^{*}$ due to tiny displacement for  ${_\mathcal{C}^{\mathcal{G}}}{{\mathbf{t}}}_{1, 2}$ in practice, where $\mathbf{v}_{i}$ represents the $i^{\text{th}}$ element of a vector $\mathbf{v}$.  

Noted that we use batch factor graph optimization  to obtain the camera-to-ground transformation matrix. Such matrix accuracy can be easily influenced by road geometry shape. To eliminate the consequence, we use rotation averaging to find the optimal rotation matrix ${_\mathcal{C}^{\mathcal{G}}} {\mathbf{R}}$ through
$
{\min} \sum_{k = 1}^{N_r} \Vert {_\mathcal{C}^{\mathcal{G}}} {\mathbf{R}}_k^{*} - {_\mathcal{C}^{\mathcal{G}}} {\mathbf{R}}\Vert_{F},
$
where $\Vert \cdot \Vert_{F}$ is the Frobenius matrix form, and $N_r$ is the moving window size (see Box 4.2 of Fig. \ref{Fig:SystemDiagram}).  A closed-form solution is given by ${_\mathcal{C}^{\mathcal{G}}} {\mathbf{R}} = \mathbf{U} \mathbf{\Sigma}  \mathbf{V}^{\intercal}$, where $\mathbf{U}$ and $\mathbf{V}$ are from SVD decomposition of $\sum_{k = 1}^{N_r} {_\mathcal{C}^{\mathcal{G}}} {\mathbf{R}}_k^{*} = \mathbf{U} \mathbf{S} \mathbf{V}^{\intercal}$. If $\mbox{det} (\mathbf{U}\mathbf{V}^{\intercal}) \ge 0$, we have $\mathbf{\Sigma} = \mbox{diag}(1, 1, -1)$. Otherwise, we have $\mathbf{\Sigma} = \mathbf{I}_{3 \times 3}$ \cite{hartley2013rotation}. Also, we estimate the translation vector by  ${_\mathcal{C}^{\mathcal{G}}} {\mathbf{t}}  = \sum_{k = 1}^{N_r} {_\mathcal{C}^{\mathcal{G}}}\mathbf{t}_k^{*}/N_r $. 
We further propose the following hypothesis testing to determine when to report/broadcast $\boldsymbol{\xi} = [\angle{{_\mathcal{C}^{\mathcal{G}}}\mathbf{R}} \; {_\mathcal{C}^{\mathcal{G}}}\mathbf{t}]^{\intercal} \in \mathbb{R}^6$ through the Z-test,
\begin{equation}\label{eq:z_test}
    \begin{aligned}
        \mathbf{H}_0: \boldsymbol{\xi} = \boldsymbol{\xi}_d,\\
        \mathbf{H}_1: \boldsymbol{\xi} \neq \boldsymbol{\xi}_d.
    \end{aligned}
\end{equation}
Here, $\boldsymbol{\xi}_d$ is the  threshold vector determined by the experiment.  
The test statistic can be calculated by 
$
    \mathbf{z} = ({\boldsymbol{\xi} - \boldsymbol{\xi}_d})/{\sqrt{\mathbf{S}/N_h}}.
$
Here, $\mathbf{S}$ is the sample covariance matrix and $N_h$ is the sample size.  
Define $\Phi(x)$ to be the cumulative distribution function of the standard normal distribution at value $x$. By setting the significance level $\alpha$, the $p$-value is obtained by $\Phi^{-1}(1 - \alpha/2)$. We report ${_\mathcal{C}^{\mathcal{G}}}\mathbf{T}$ by failing to reject $\mathbf{H}_0$ when
$
\vert \mathbf{z} \vert \prec \Phi^{-1}(1 - \alpha/2).
$
Otherwise, we continue our cross-keyframe ground refinement process until the alternative hypothesis $\mathbf{H}_1$ is rejected.

\subsection{Failure Detection and Recovery}
Although our system is robust to various driving conditions and road geometry shapes, failure is still unavoidable due to severe motion or illumination change.  We use active failure detection and recovery strategy to improve the practicability of the proposed system. Failure
detection is an independent module that detects unusual output from the estimator. We are currently using the following
criteria for failure detection:  large discontinuity from the rotation or position between two keyframes; the number of tracked ground features in the new keyframe is less than a threshold; ground normal vector and camera center-to-ground height estimation have large changes; the number of triangulated points fails to provide a minimum solution; the ground normal vector and camera center-to-ground height fail our quality tests. 
Once a failure is detected, the system switches back to reinitialize the system, and restart to accumulate observations for a new and separate segment of the factor graph. 

\begin{table}[t]
	\caption{Results on FPG dataset} 
	\centering  
	\begin{tabular}[htb!]{ c c c c c }
		\hline
		\\\\[-2.5\medskipamount]
		  Method & $\delta_{r}^{\degree}$ & $\delta_{p}^{\degree}$ & $\delta_{y}^{\degree}$  & $\delta_{h} (cm)$ \\ 
		\\\\[-2.8\medskipamount]
		\hline	 		
		\\\\[-2.8\medskipamount]
            {ROECS}        & 0.14   & 0.18  & 0.30  &  1.70 \\ 
            {OECS}         & 0.54   & 0.20  & 0.22  &  2.10 \\ 
		{Liu \textit{et al.}}   & 0.10   & 0.09  & 0.31  & 0.59 \\ 	
		{\textbf{Ours}}         & \textbf{0.10}  & \textbf{0.05}  & \textbf{0.11}  & \textbf{0.17} \\ 
		\\\\[-2.5\medskipamount] 
		\hline
	\end{tabular} \label{tab::FPGresults}
	\vspace{-6mm}
\end{table}

\section{Experiment}\label{sec:experiment}

We implement our algorithm and perform extensive experiments under various of driving scenarios using full-size passenger cars.  The vehicles are installed with surround view camera systems made of four downward-facing fisheye cameras (see samples in Fig. \ref{Fig:FEimages}), which are synchronized with wheel odometry data. The cameras run at $33$ Hz, and the image resolution is resized to be $812$$\times$$540$. 
We collect long-sequence of continuous data from different areas  to analyze our algorithm's efficiency and robustness (see Tab. \ref{tab::datainfo}). 
They vary  from a flat paved ground (FPG) to public road such as city, suburban and rural regions under different weather, lighting, and driving conditions. The FPG data are from  extremely flat asphalt surfaces, which are used to verify baseline performance of our dynamic calibration.  The last column of Tab. \ref{tab::datainfo} indicates the percentage of time when the vehicle is moving instead of parking.

\begin{figure}[t]
    \centering
    \subfigure[Texture-less Env.]{\includegraphics[width=1in]{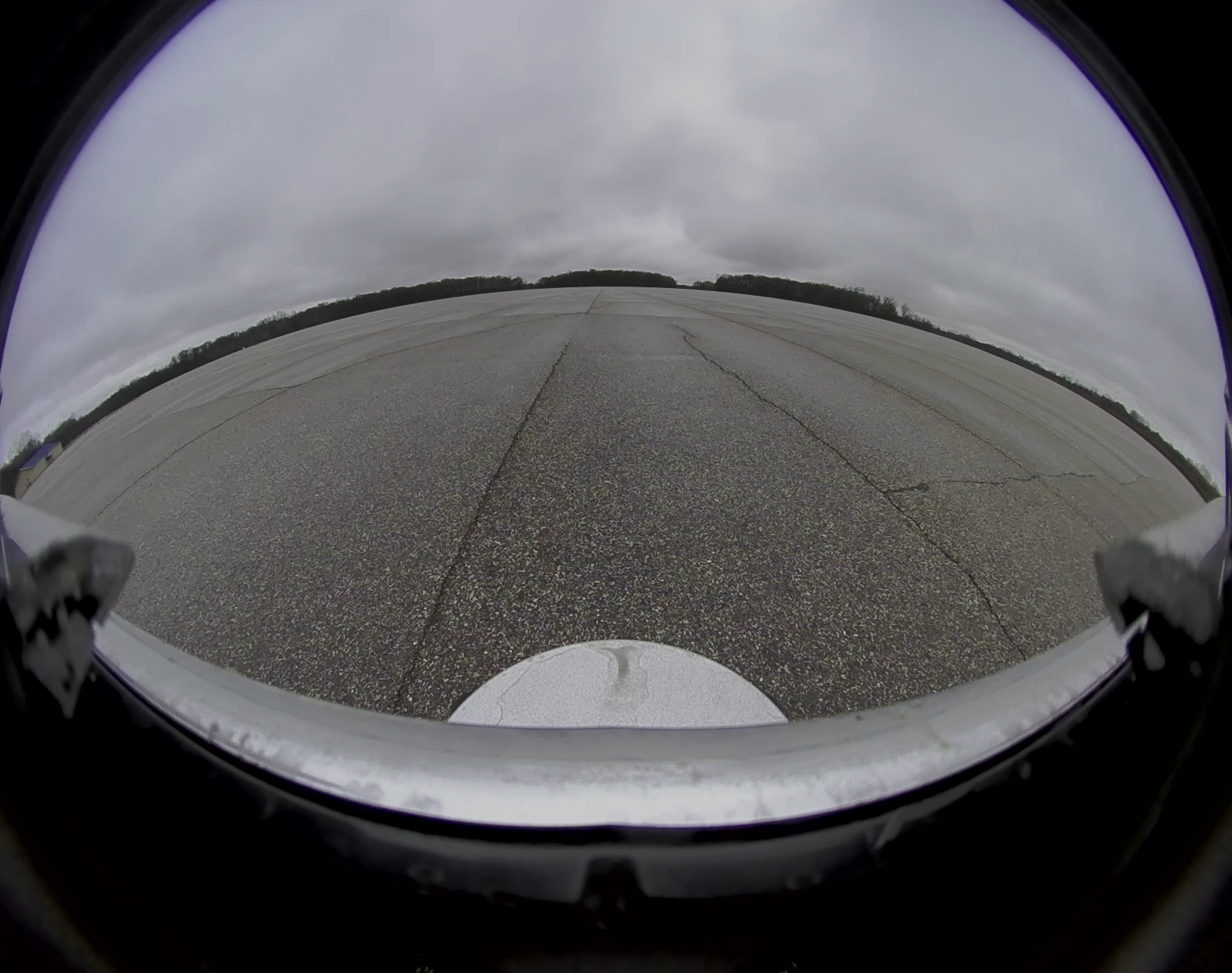}\label{Fig:textureless}}
    \subfigure[City Driving]{\includegraphics[width=1in]{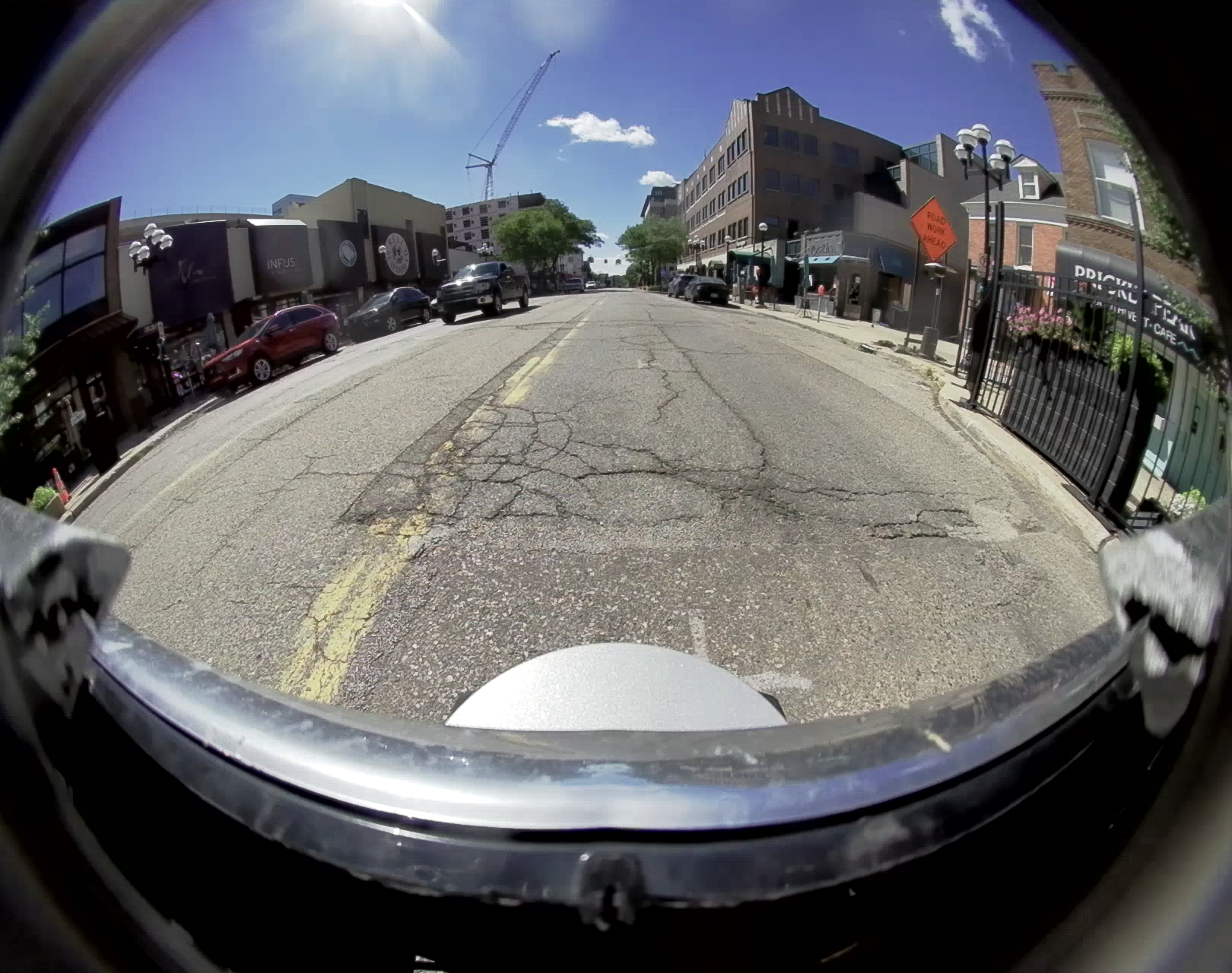}\label{Fig:cities}}
    \subfigure[Night Driving]{\includegraphics[width=1in]{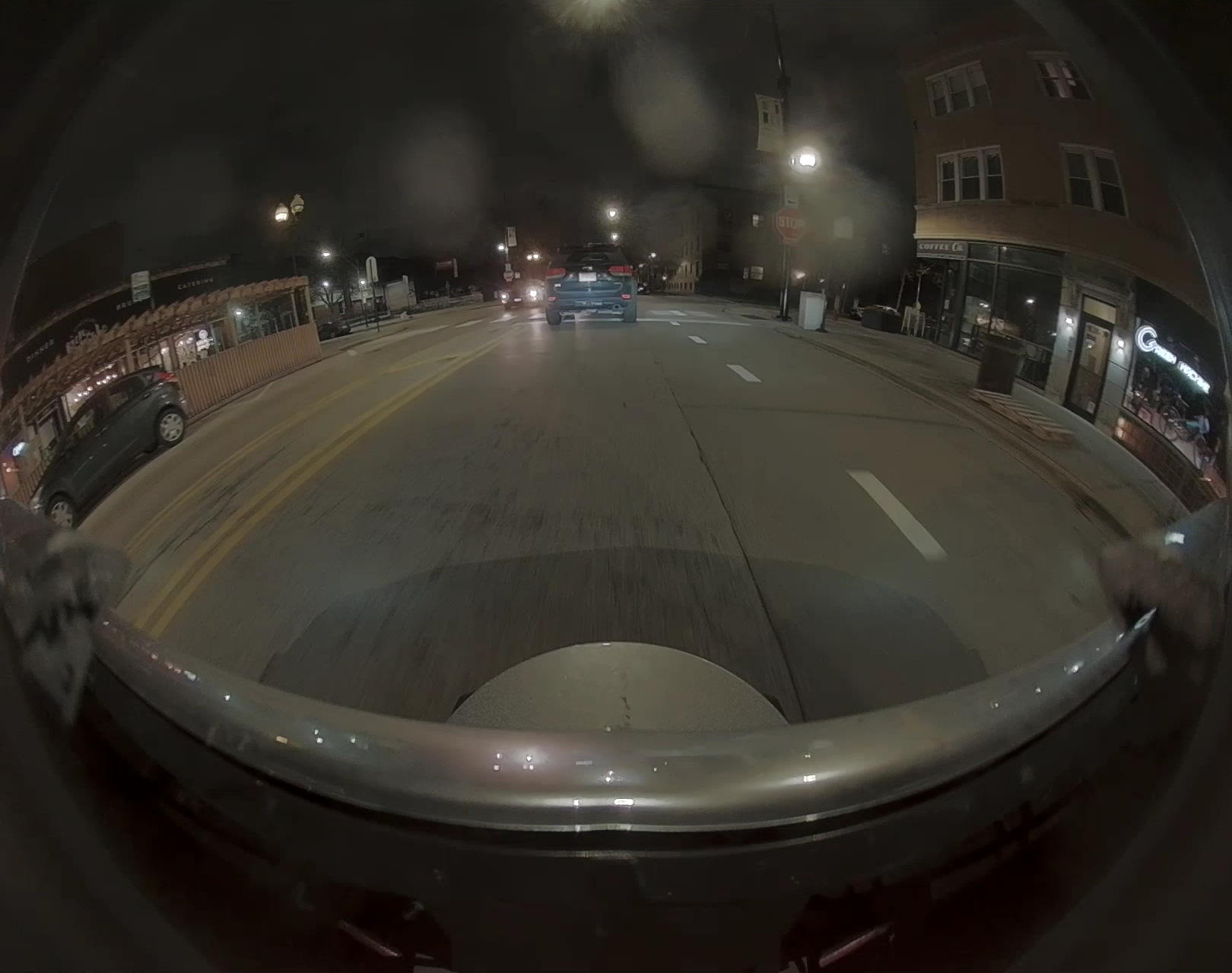}\label{Fig:nightcity}}
    \subfigure[Crowded Traffic]{\includegraphics[width=1in]{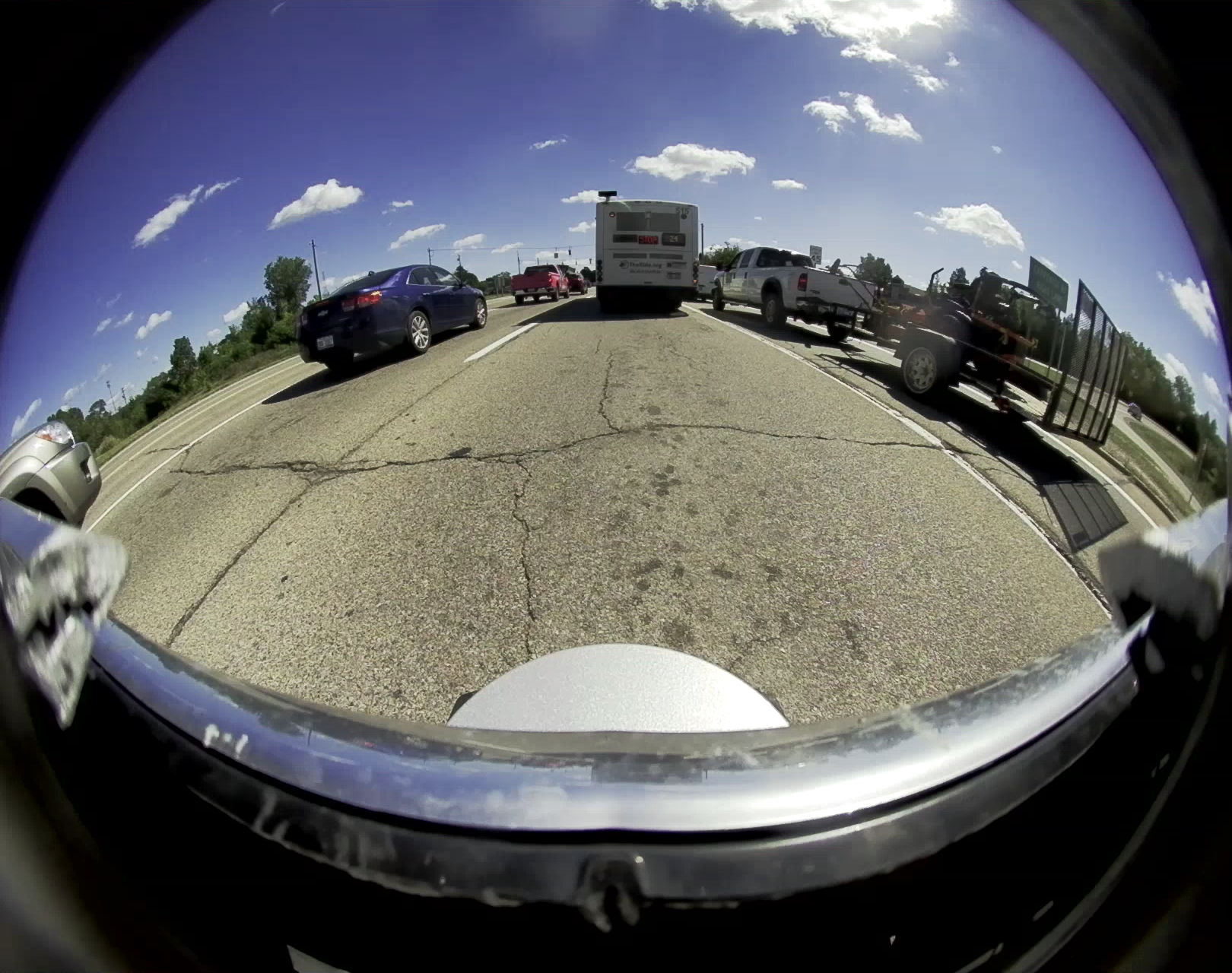}\label{Fig:crowdedtraffic}}
    \subfigure[Rainy Weather]{\includegraphics[width=1in]{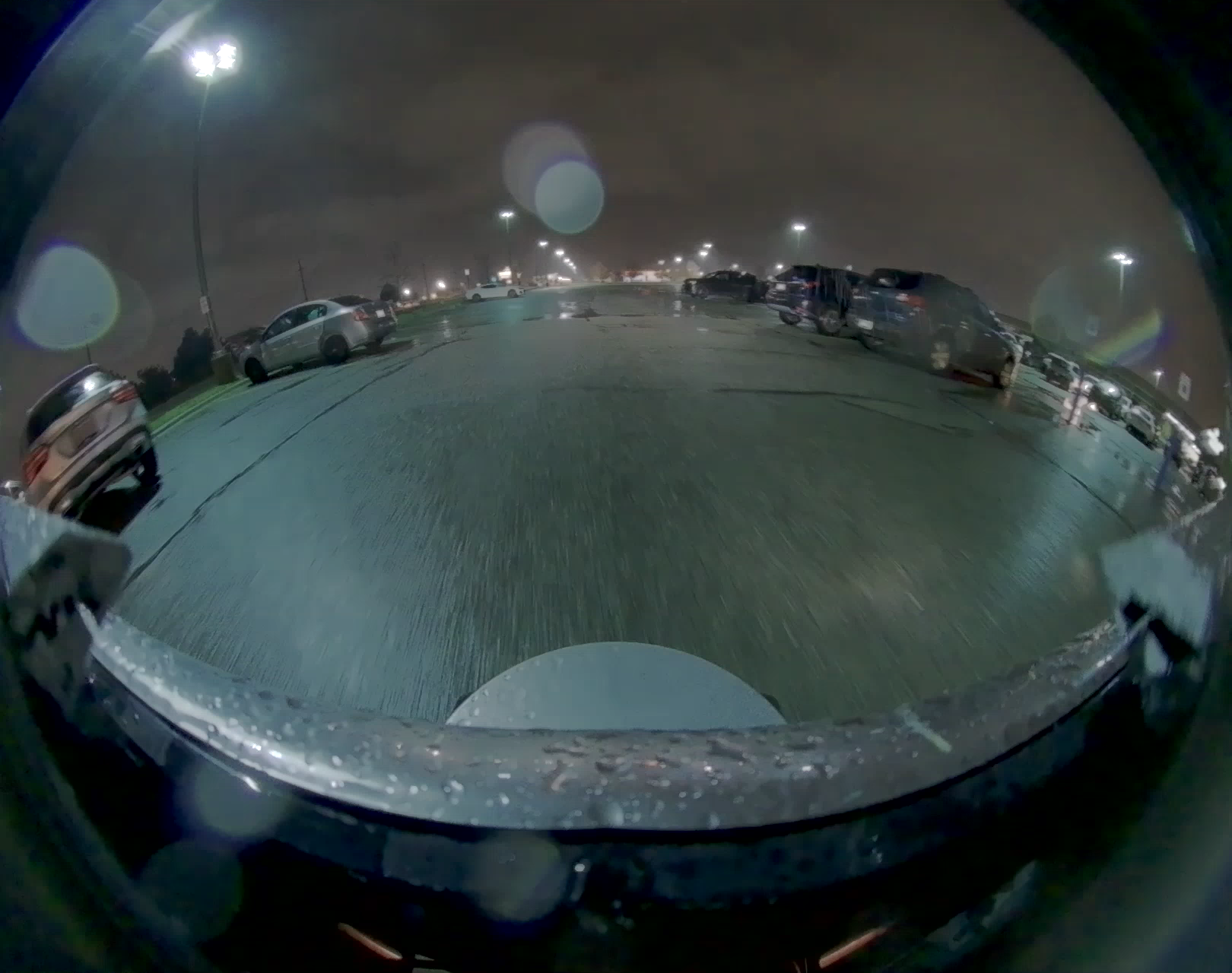}\label{Fig:rain}}
    \subfigure[Rural Road]{\includegraphics[width=1in]{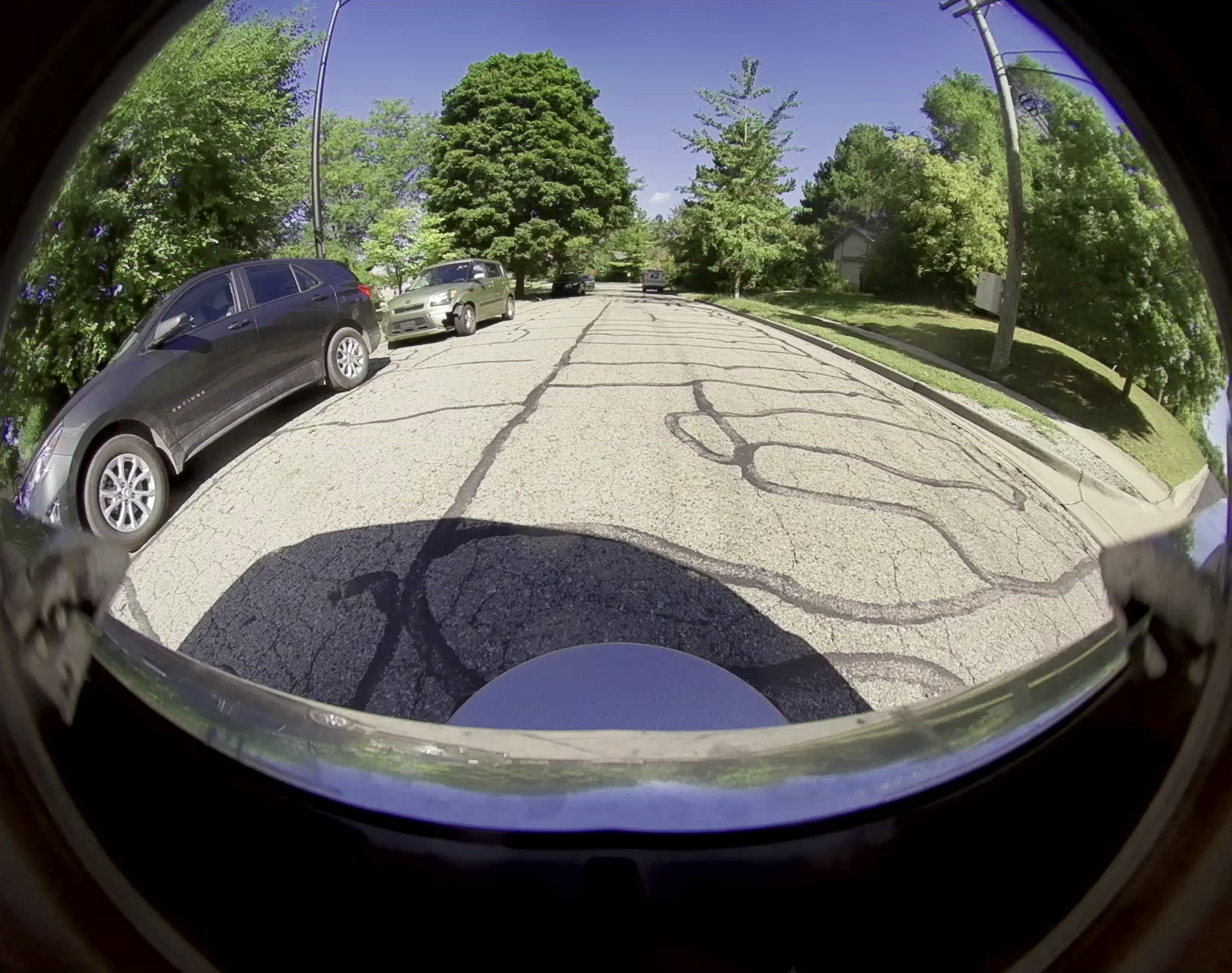}\label{Fig:nolanemarkings}}	
    \caption{Our approach performs well in challenging  driving scenarios. Figure labels correspond Tab. \ref{tab::datainfo} from the top to the bottom. 
    }\label{Fig:various_scenarios}
    \vspace{-5mm}
\end{figure}

We compare our approach with the state-of-the-art approaches to qualitatively examine the performance, including Liu \textit{et al.} \cite{liu2019online}, OECS \cite{zhang2020oecs}, and ROECS \cite{zhang2021roecs} on our dataset.
We first compare the calibration performance on the FPG data  (sample pictures in Fig. \ref{Fig:textureless}), and summarize results in Tab. \ref{tab::FPGresults}.
Like our counterparts, our approach generates continuous and stable camera-to-ground calibration while the vehicle is on FPG. Thus we show the average of the Euler angle difference with the ground-truth (GT) calibration, and evaluate the absolute changes as $\delta_{r}$,  $\delta_{p}$,  and $\delta_{y}$. Here,  $\delta_{r}$,  $\delta_{p}$,  and $\delta_{y}$ are the roll, pitch and yaw angle changes, respectively. The column $\delta_{h}$ in Tab. \ref{tab::FPGresults} is the camera center-to-ground height displacement. Noted that we obtain the GT calibration by using a highly accurate positioning system and it is performed offline by refining the vehicle CAD values.  
In Tab. \ref{tab::FPGresults},  our online approaches achieve better performances even on concrete road surface without distinguishable textures on images. For instance, our methods obtain significant absolute gains on pitch, yaw, and height estimation by $44.4\%$, $50.0\%$ and $71.2\%$, respectively.

Due to the lack of the GT while driving on public road, we propose metrics to measure the accuracy of the extrinsics estimated by the compared methods. Define $\mathbf{F}_\mathcal{Q}$ to be the fundamental matrix between camera $\mathcal{Q}$ and the front-facing camera, where $\mathbf{F}_\mathcal{Q} = ({\mathbf{K}_{\mathcal{Q}}^{-1}})^{\intercal}\eqspace\lfloor {_{\mathcal{Q}}^{\mathcal{F}} \mathbf{t}} \rfloor_{\times} \eqspace{_{\mathcal{Q}}^{\mathcal{F}}\mathbf{ R}} \eqspace{\mathbf{K}_{\mathcal{F}}^{-1}}$, and $\mathbf{K}_{\mathcal{X}}$ is the intrinsic matrix of a camera $\mathcal{X} \in \{ \mathcal{L}, \mathcal{R}, \mathcal{F} \}$. Here, ${_{\mathcal{Q}}^{\mathcal{F}}\mathbf{T}} = {_{\mathcal{G}}^{\mathcal{F}}\mathbf{T}} {_{\mathcal{G}}^{\mathcal{Q}}\mathbf{T}^{-1}}$, and $\mathcal{L}$,  $\mathcal{R}$ and $\mathcal{F}$ represent the left-facing, right-facing and front-facing cameras, respectively.
We have the following, 
\begin{itemize}
    \item \textbf{Feature transfer error}: We utilize (\ref{eq:H}) and (\ref{eq:optimize}) to obtain the homography matrix between neighbouring keyframes and propose to utilize,
    \begin{equation}\label{eq:transfer_error}
       \epsilon_{f} = \frac{1}{N_s}\sum_{j = 1}^{N_s} \Vert \mathbf{r}_{k, j}\Vert, 
    \end{equation}
    to quantify the extrinsic qualities within a single camera. Here, $N_s$ is the set cardinality of  $\mathcal{Q}_{k \leftrightarrow k +1}$,  and $\mathbf{r}_{k, j}$ is the feature distance in (\ref{eq:optimize}). 
    \item \textbf{Feature residual error}:  We leverage the squared distance  between a feature point’s epipolar line and its matching point  in the other camera by averaging over all $N_f$ matches, 
    \begin{equation}\label{eq:feature_error}
       \epsilon_{p} = \frac{1}{\vert N_f \vert} \sum_{l = 1}^{N_f} d(\mathbf{q}_{k, l}, {\mathbf{F}_\mathcal{Q}}\; {\check{\mathbf{p}}}_{k, l})^2 + d(\mathbf{p}_{k, l}, {\mathbf{F}_{\mathcal{Q}}}^\intercal \; {\check{\mathbf{q}}}_{k, l})^2,
    \end{equation}
    to measure calibration performance across different cameras, where $\mathbf{p}_{k, l} \in \mathcal{F}$ and $\mathbf{q}_{k, l} \in \mathcal{Q}$ are matched features, 
    and $d(\mathbf{p}, \mathbf{l})$ is the distance of a point $\mathbf{p}$ to the line $\mathbf{l}$. 
\end{itemize}
Smaller errors are preferred for the abovementioned metrics. 
For brevity, we only use overlapping regions of  the front-facing camera with the left-facing and right-facing cameras. Other overlapping regions of surround view camera systems can be integrated in (\ref{eq:feature_error}) as well.  

\begin{figure}[tb]
    \centering
    \subfigure[FPG]{\includegraphics[width=1.6in, viewport = 5 0 420 310, clip]{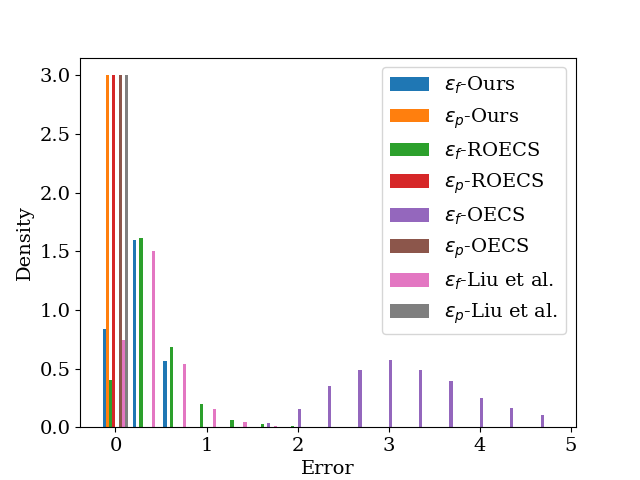}\label{Fig:FPG_error}}
    \subfigure[City I]{\includegraphics[width=1.6in, viewport = 5 0 420 310, clip]{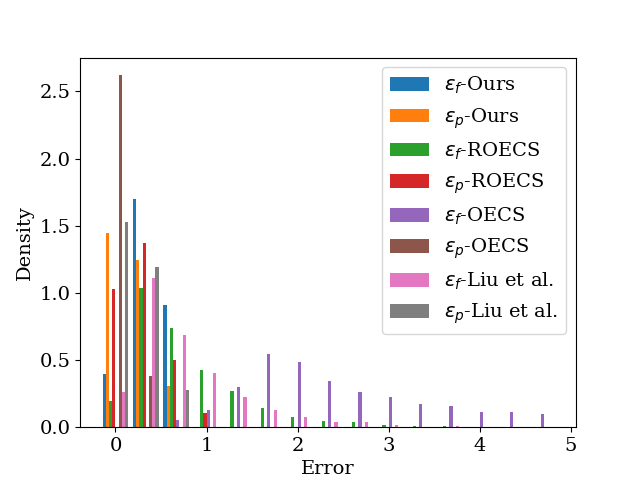}\label{Fig:City1}}
    \subfigure[City II]{\includegraphics[width=1.6in, viewport = 5 0 420 310, clip]{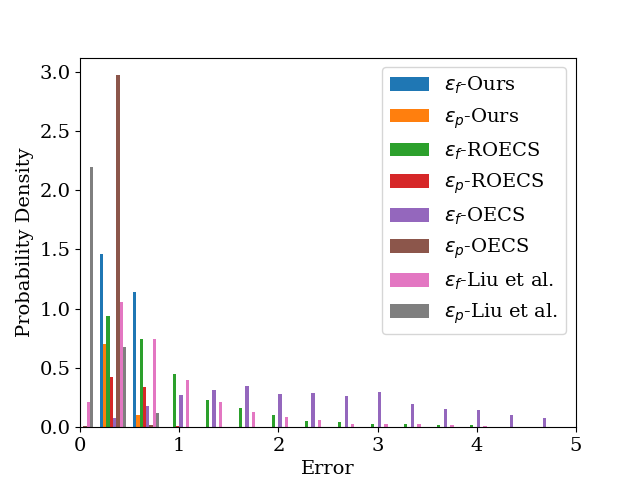}\label{Fig:City2}}
    \subfigure[Urban]{\includegraphics[width=1.6in, viewport = 5 0 420 310, clip]{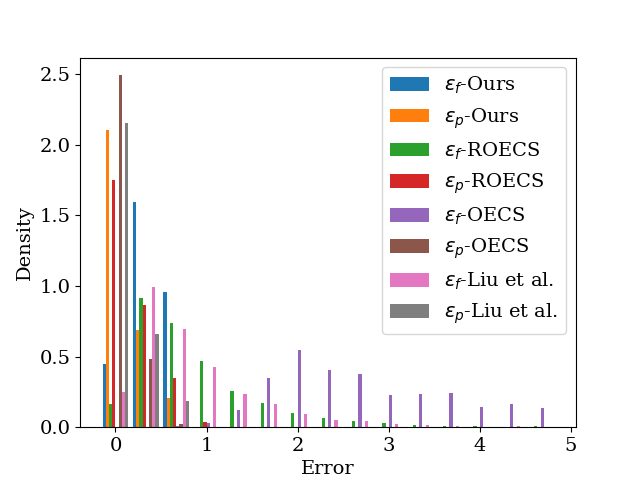}\label{Fig:Urban}}
    \subfigure[Suburban]{\includegraphics[width=1.6in, viewport = 5 0 420 310, clip]{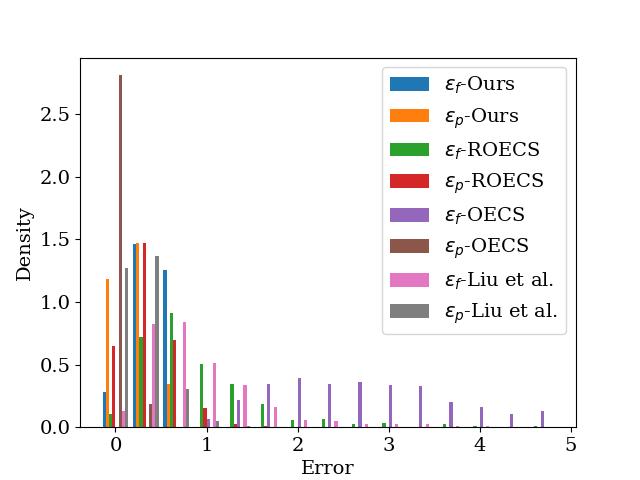}\label{Fig:subUrban}}
    \subfigure[Rural]{\includegraphics[width=1.6in, viewport = 5 0 420 310, clip]{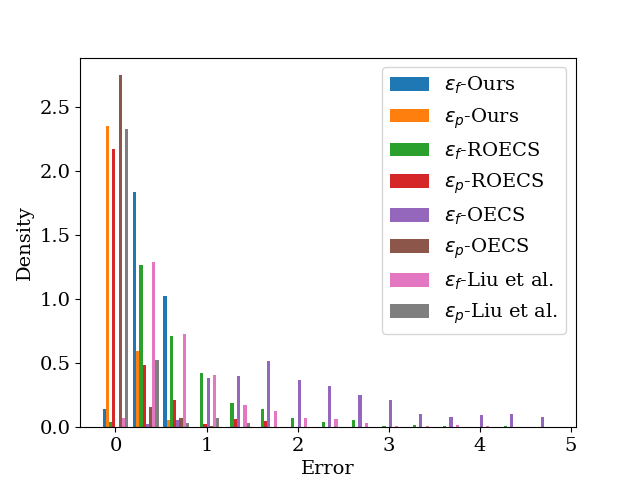}\label{Fig:Rural}}	
    \caption{The histogram graphs of performance for data sequences in Tab. \ref{tab::datainfo}. The value of the horizontal axis corresponds to errors in (\ref{eq:transfer_error}) and (\ref{eq:feature_error}), and the vertical axis is the probability density. \textit{Zoom in for details.} }\label{Fig:Errors}
    \vspace{-7mm}
\end{figure}

To demonstrate the superiority of our proposed methods, we present the histogram plots of errors of (\ref{eq:transfer_error}) and (\ref{eq:feature_error}) in Fig. \ref{Fig:Errors}. 
We also obtain the error differences at different bins in Fig. \ref{Fig:Errors} to have the summation as our performance gain over the state-of-art work. 
Our approach has a relatively smaller feature transfer error $\epsilon_f$ under different levels and the error is maintained within $0.83$ pixels, which is consistent across different data sequences. 
Our method achieves  the lowest performance gain $44.2\%$ on the Suburban data, and the highest performance gain $67.6\%$ on the Urban data.   
On public road driving data ranging from city to rural area, we achieve better performance than the state-of-the-art methods when comparing the error $\epsilon_p$. 
The error $\epsilon_p$ varies within $0.75$ pixels, and $90.9\%$ of errors are within 0.67 pixels by our methods. 
The performance is improved by $12.7\%$ with the highest score on the  Suburban data, and the lowest score of $1.59\%$ on the FPG data considering  flat road conditions. 

\begin{figure}[tb]
    \centering
    \subfigure[]{\includegraphics[height=1.2in, width = 0.8in]{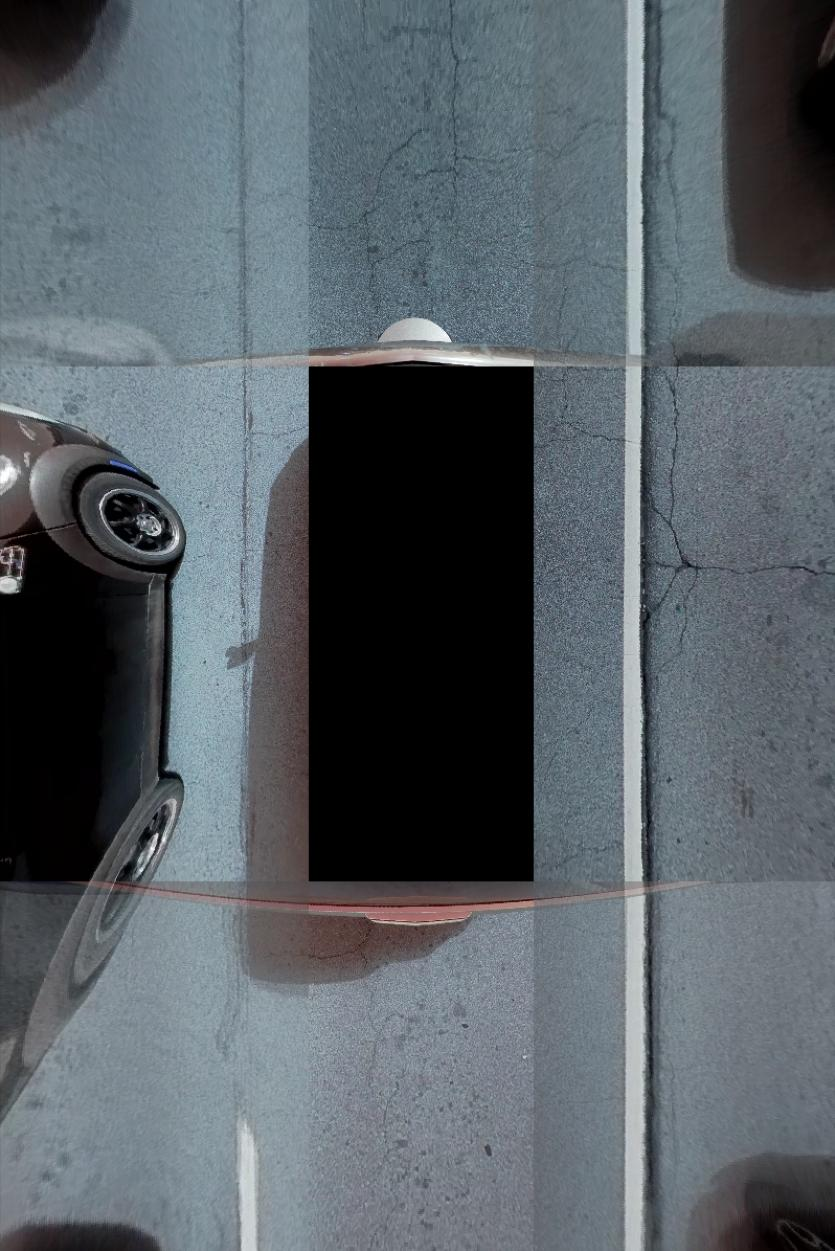}\label{Fig:figure1}}
    \subfigure[]{\includegraphics[height=1.2in, width = 0.8in]{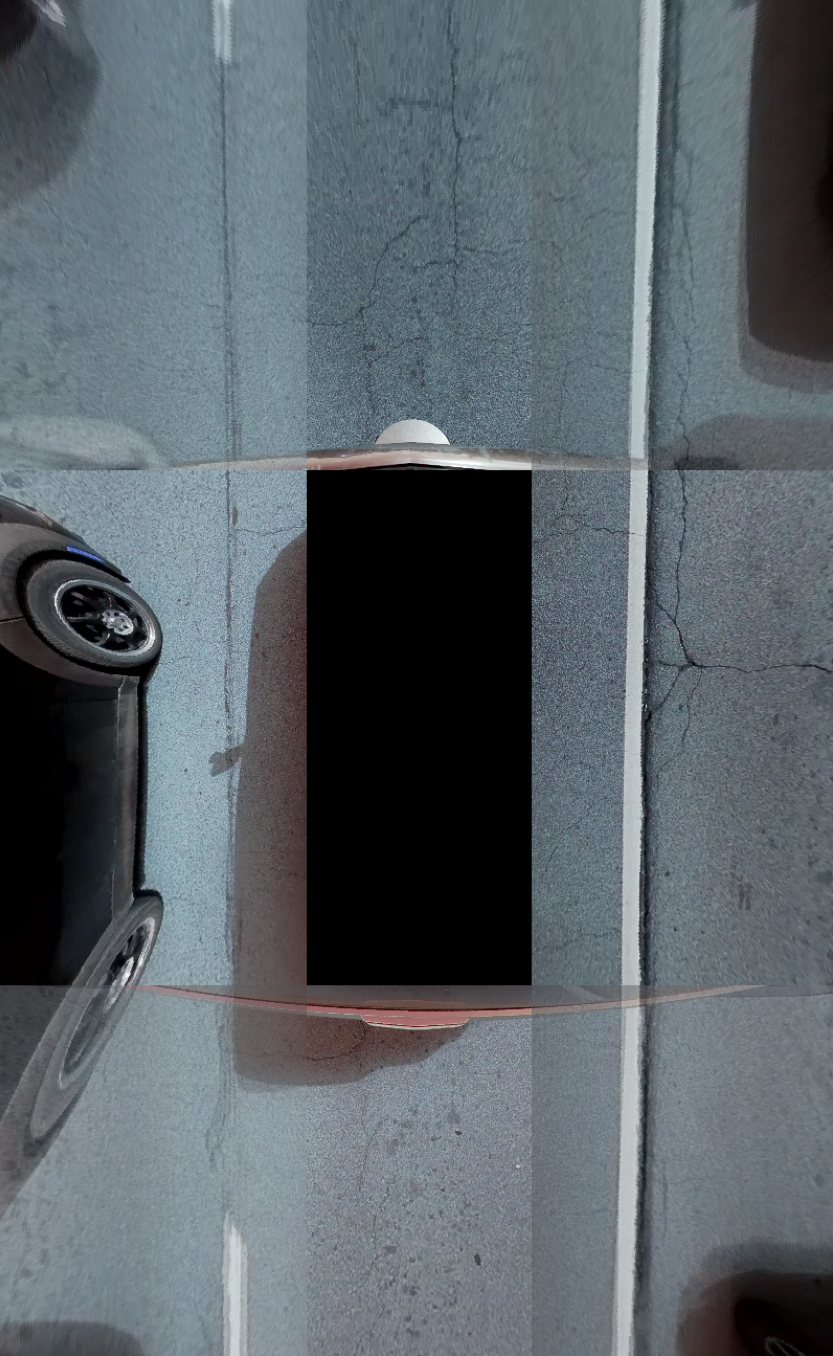}\label{Fig:figure2}}
    \subfigure[]{\includegraphics[height=1.2in, width = 0.8in]{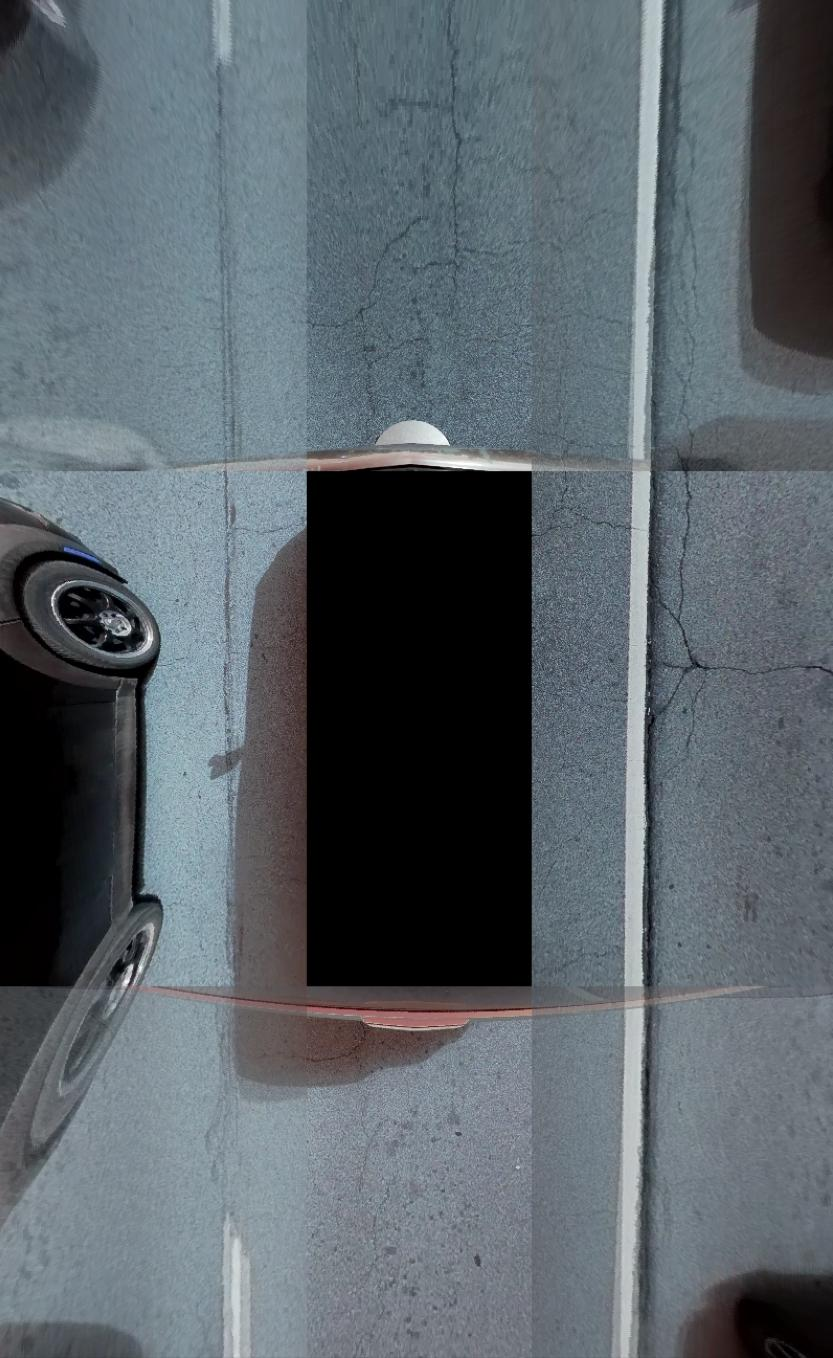}\label{Fig:figure3}}
    \subfigure[]{\includegraphics[height=1.2in, width = 0.8in]{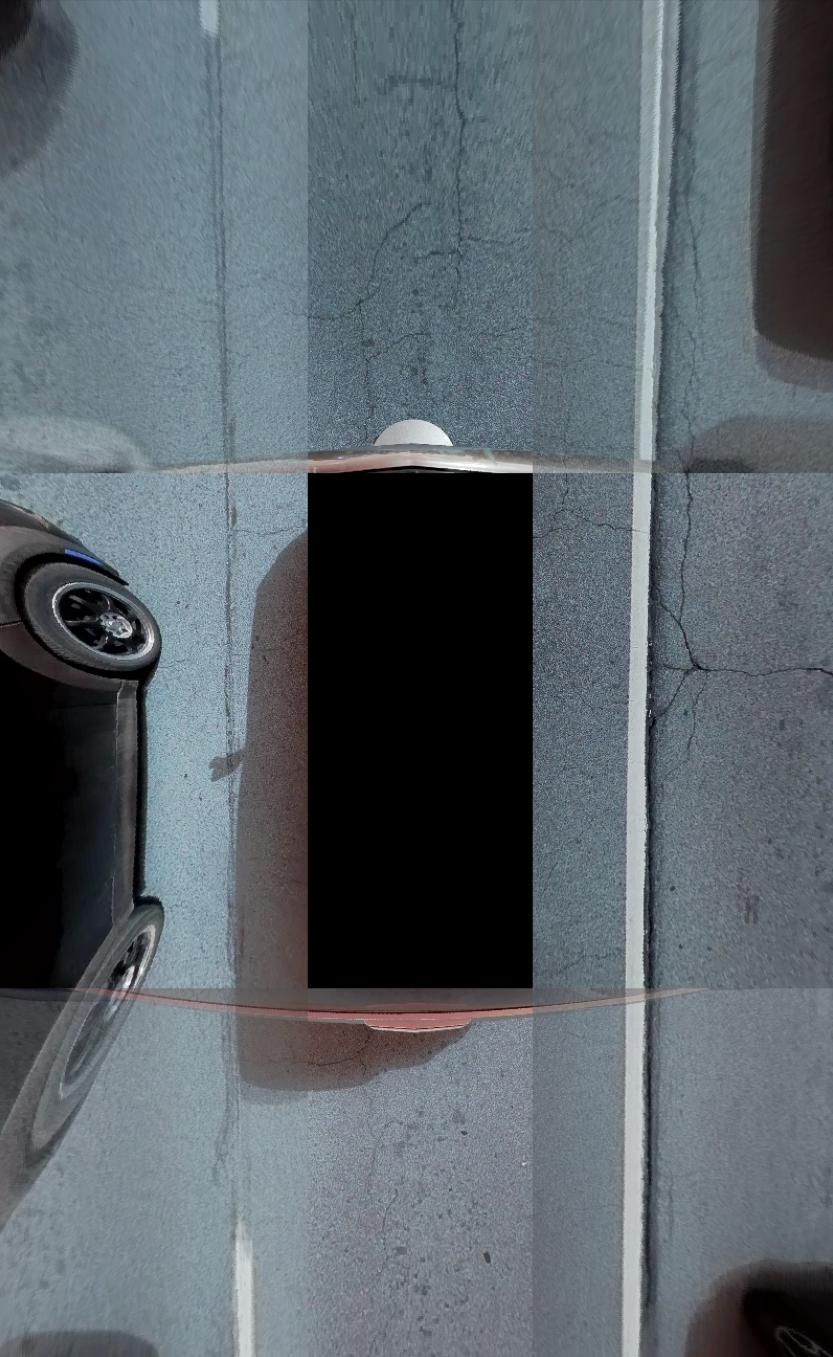}\label{Fig:figure4}}
    \caption{Example results on City I data while driving. We present (a) our result, (b)  ROECS \cite{zhang2021roecs}, (c)  OECS \cite{zhang2020oecs}, and (d) Liu \textit{et al.} \cite{liu2019online} given ${_\mathcal{C}^{\mathcal{G}}}\mathbf{T}$ from different approaches. \textit{Zoom in for better viewing.} 
    }\label{Fig:sample_scenarios}
    \vspace{-7mm}
\end{figure}

Fig. \ref{Fig:sample_scenarios} shows visual results of the BEV images from different methods when the vehicle is known to be driving straight at a high speed. Without relying on the overlapping regions or specific objects across different cameras, our approach obtains smaller feature residual errors and generates a better-aligned BEV image given the facts: (1) straight lane markings are parallel to the vehicle's driving direction, (2) lane markings across cameras overlap each other, and (3) concrete cracks on the road are connected across the overlapping regions of neighboring camera's BEV views.

\section{Conclusion and Future Work}\label{sec:conc}

We proposed an online camera-to-ground targetless calibration method to generate a non-rigid body transformation between the camera coordinate and the ground coordinate while driving. We utilized a novel coarse-to-fine architecture to select ground features and  verified them through a geometry-based approach.
We performed plane fitting for triangulated ground features to attain ground normal vectors and camera center-to-ground height, which were refined through factor graph optimization in a sliding window.
We determined the camera-to-ground transformation through rotation averaging and provided stopping criteria to report/broadcast satisfying calibration results.
We extensively tested our algorithm with real data collected from different weather and driving conditions. The results showed that our method is effective and outperforms state-of-the-art techniques.

In the future, we will reduce the running time complexity for factor graph optimization and perform observability analysis to identify the degenerate motion segments that help to discard poses and ground features which are not necessary for calibration computation. 

\section*{Acknowledgment}
\small{The author would like to thank A. Kurup, G. He, I. Hamieh, M. Khalili, S. Gagnon, X. Guo, X. Liu, A. Farah, S. Miller, Y. Zhang and L. Li for their great support to this research.}

\bibliographystyle{IEEEtran}
\bibliography{./li.bib}

\end{document}